\pdfoutput=1

\documentclass[11pt]{article}

\usepackage[preprint]{acl}

\usepackage{times}
\usepackage{latexsym}

\usepackage[T1]{fontenc}

\usepackage[utf8]{inputenc}

\usepackage{microtype}

\usepackage{inconsolata}

\usepackage{graphicx}

\usepackage{booktabs}
\usepackage{multirow}
\usepackage{subfigure}
\usepackage{amsmath}
\usepackage{amssymb}
\usepackage{float}

\newcommand{\ourbtmodel}{Rc-BT}
\newcommand{\ourrm}{Rc-RM}
\newcommand{\ourdpo}{Rc-DPO}
\newcommand{\QEA}{{\em Quality Eval Acc}}
\newcommand{\LEA}{{\em Length Eval Acc}}

\title{Disentangling Length Bias in Preference Learning via Response-Conditioned Modeling}

\author{
  Jianfeng Cai\footnotemark[2] \And Jinhua Zhu\thanks{Corresponding author.}\footnotemark[2] \And Ruopei Sun\footnotemark[2] \\
  \AND Yue Wang\footnotemark[3] \And Li Li\footnotemark[2] \And Wengang Zhou\footnotemark[2] \And Houqiang Li\footnotemark[2] \\
  \AND \textnormal{\footnotemark[2]\,\,University of Science and Technology of China} \\ \textnormal{\footnotemark[3]\,\,Independent Researcher} \\
  \texttt{\{xiaobaicai,teslazhu\}@mail.ustc.edu.cn}
}

\begin{document}
\maketitle

\begin{abstract}
Reinforcement Learning from Human Feedback~(RLHF) has achieved considerable success in aligning large language models~(LLMs) by modeling human preferences with a learnable reward model and employing a reinforcement learning algorithm to maximize the reward model's scores. However, these reward models are susceptible to exploitation through various superficial confounding factors, with length bias emerging as a particularly significant concern. Moreover, while the pronounced impact of length bias on preference modeling suggests that LLMs possess an inherent sensitivity to length perception, our preliminary investigations reveal that fine-tuned LLMs consistently struggle to adhere to explicit length instructions. To address these two limitations, we propose a novel framework wherein the reward model explicitly differentiates between human semantic preferences and response length requirements. Specifically, we introduce a \textbf{R}esponse-\textbf{c}onditioned \textbf{B}radley-\textbf{T}erry~(Rc-BT) model that enhances the model's capability in length bias mitigating and length instruction following, through training on our augmented dataset. Furthermore, we propose the Rc-RM and Rc-DPO algorithm to leverage the Rc-BT model for reward modeling and direct policy optimization~(DPO) of LLMs, simultaneously mitigating length bias and promoting adherence to length instructions. Extensive experiments across various foundational models and datasets demonstrate the effectiveness and generalizability of our approach.
\end{abstract}
\section{Introduction}
\label{sec:introduction}
Reinforcement learning from human feedback (RLHF)~\citep{ziegler2019fine, stiennon2020learning, ouyang2022training} has revolutionized the field of natural language processing, enabling advancements in areas such as conversation~\citep{bai2022training}, code generation~\citep{poesia2022synchromesh, 10.1145/3672456}, complex planning~\citep{hao-etal-2023-reasoning, zhao2024large}, mathematical reasoning~\citep{imani2023mathprompter, luo2023wizardmath}, and so on.
Within this framework, preference learning plays a pivotal role by using a generalizable model as a proxy for human preferences and optimizing large language models~(LLMs) based on this preference model.
However, a significant challenge in RLHF is the phenomenon termed reward overoptimization or hacking~\citep{gao2023scaling}.
This occurs when excessive optimization against the preference model undermines the attainment of the true objective.
As a result, LLMs may learn to exploit simpler criteria such as length, bullet points, or politeness~\citep{rame2024warm}, rather than more causal and nuanced indicators to achieve higher reward.

Among these reward hacking phenomena, length bias stands out as both a prevalent and challenging issues~\citep{dubois2024length}, and we identify it as a representative test bed, with the expectation that our approach can be readily extended to address other spurious correlations.
To mitigate length bias in the RLHF pipeline, two primary strategies have emerged, which may be complementary: The first strategy focuses on rectifying the policy optimization process through techniques such as increasing the KL penalty coefficient, applying length penalty rewards~\citep{singhal2023long}, reward clipping~\citep{chenodin}, and various hyper-parameter tuning methods.
The second strategy aims to disentangle length information from quality in reward modeling.
This includes length balancing, reward data augmentation, confidence-based truncation in training data~\citep{singhal2023long}, and using sophisticated training objectives to separate length information~\citep{chenodin}.

Despite these advances, we observe several drawbacks in these methods.
First, rectification methods for policy optimization are sensitive to hyperparameter changes and foundational model variations, and prior work demonstrates that their effectiveness is limited~\citep{singhal2023long, chenodin}.
The second category shows superior performance compared to the first; however, over-parameterization of the two branches can cause unstable optimization~\citep{kurach2019the}, and regularization methods that encourage linear irrelevance may not guarantee true independence~\citep{shimony1993role}.
Furthermore, ignoring the biased nature of the training data and forcing an irrelevancy or orthogonality between response quality and response length is not always reasonable. For instance, in some instruction-following datasets, such as length instruction dataset, response length may indeed be a factor of response quality.
Finally, these methods regard length information as harmful and aim to minimize its influence on the instruction-following ability. However, whether we could leverage these length biases for better preference modeling remains an underexplored question.

Based on the aforementioned observations, we aim to improve preference modeling by leveraging response length information
and propose a novel approach that allows the preference model to explicitly differentiate between human semantic intent and response length instructions. Specifically, we first develop an augmented length instruction dataset derived from the original preference dataset. Subsequently, we introduce a Response-conditioned Bradley-Terry (\ourbtmodel{}) model, which improves the model's ability to mitigate length bias and follow length instructions.

Given the widespread use of paired data preference modeling in RLHF, our method can be seamlessly integrated into both the reward modeling and policy optimization stages. First, we empirically verify our method in the reward modeling task. The results demonstrate that our method not only enhances the reward model's ability to mitigate length bias, thereby improving its alignment with human semantic quality, but also strengthens its adherence to length instructions. Furthermore, applying our method to direct preference optimization (DPO) \citep{rafailov2024direct}, a popular alignment algorithm, further validates its effectiveness.

Our contributions are summarized as follows: (1) We propose the Response-conditioned Bradley-Terry method to mitigate length bias and enhance the model's capacity to follow length instructions. (2) We show that our method can be integrated into reward modeling and policy optimization with minimal adjustments. (3) Experimental results illustrate the superiority and generalizability of our method across multiple models and datasets.
\section{Related Work}
\label{sec:related-work}

In this section, we briefly introduce the background of RLHF, reward hacking and length instruction following. A more detailed version is in Appendix~\ref{apx:full-related work}.

{\bf Reinforcement Learning From Human Feedback.} RLHF~\citep{ziegler2019fine} involves training a reward model to approximate human preferences and optimizing the LLMs through reinforcement learning (RL)~\citep{schulman2017proximal}. However, this approach suffers from training instability and requires careful tuning of numerous hyperparameters. Recent direct preference alignment methods, particularly DPO~\citep{rafailov2024direct}, offer a more stable alternative. By reformulating the reward function, DPO eliminates the need for an online reward model, enabling robust offline preference learning~\citep{hong2024orpo, chen2024noise, ethayarajh2024kto}. Our method not only enhances RLHF by improving reward model but also seamlessly integrates into the DPO framework.

{\bf Reward Hacking.} RLHF is vulnerable to reward hacking, where LLMs exploit inherent biases in the proxy preference model to achieve higher scores~\citep{pan2022effects, casper2023open, lambert2023alignment}. This phenomenon persists in DPO, primarily arising from task complexity, evaluation limitations, and biased feedback~\citep{dubois2024alpacafarm}. One common form of reward hacking is length bias~\citep{singhal2023long, park2024disentangling}, wherein models favor longer responses regardless of semantic quality. Previous attempts to address this issue include ODIN's~\citep{chenodin} length regularization in reward modeling; dual-model training with different learning rates~\citep{shen2023loose}; and DPO objective modification with length penalties~\citep{park2024disentangling}. In contrast to these methods that suppress length information, our method enables models to distinguish between semantic intent and length instructions, preserving both aspects in a balanced manner.

{\bf Length Instruction Following.} Current LLMs can respond to qualitative length descriptors like "concise" or "verbose" but struggle with explicit numerical constraints such as "150 words or less"~\citep{yuan2024following}. Although LIFT~\citep{yuan2024following} improves length instruction adherence through specialized datasets, this approach compromises semantic quality, resulting in degraded overall performance. Our method, in contrast, achieves effective length instruction following while simultaneously enhancing semantic quality.
\begin{table*}[!ht]
    \centering
    \vspace{-0.2cm}
    \caption{Evaluation results of reward models (Baseline) on different evaluation datasets $\mathcal{D}_{eval}$, $\mathcal{D}_{eval}^e$ and $\mathcal{D}_{eval}^r$.}
    \vspace{-0.2cm}
    \begin{tabular}{lcccc}
    \toprule
        \multirow{2}{*}{Model} & \multicolumn{3}{c}{Accuracy ($\%$)} & Consistency ($\%$) \\ \cmidrule(lr){2-4}
        & $\mathcal{D}_{eval}$ & $\mathcal{D}_{eval}^e$ & $\mathcal{D}_{eval}^r$ & $\mathcal{D}_{eval}^e$ ($\mathcal{D}_{eval}^r$) \\
        \midrule
        Qwen2-1.5B-Base & 63.86 & 64.13 & 64.40 & 89.40 (87.34) \\ 
        Qwen2-1.5B-Instruct & 62.77 & 65.22 & 60.87 & 92.12 (90.42) \\ 
        \midrule
        Qwen2.5-7B-Base & 57.07 & 60.05 & 60.33 & 88.32 (90.13) \\
        Qwen2.5-7B-Instruct & 63.04 & 62.23 & 60.60 & 88.60 (89.85) \\
        \midrule
        Llama-3.1-8B-Base & 56.25 & 60.05 & 61.41 & 89.40 (85.24) \\
        Llama-3.1-8B-Instruct & 57.88 & 58.15 & 56.25 & 88.59 (87.26) \\
        \bottomrule
    \end{tabular}
    \vspace{-0.4cm}
    \label{tab:baseline-rm-original-empty-prompt-result}
\end{table*}
\section{Preliminary Explorations}
\label{sec:preliminary}

In this section, we conduct several preliminary investigations into the ability of LLMs to perceive and process length information, with an emphasis on the reward model. Following~\citet{yuan2024following}, we utilize the OpenAssistant dataset~\citep{kopf2024openassistant}, which is partitioned into three subsets: $\mathcal{D}_{sft}$ for supervised fine-tuning (SFT), $\mathcal{D}_{rm}$ for reward model (RM) training, and $\mathcal{D}_{eval}$ for RM evaluation. Additionally, we employ a range of models for subsequent experimental analyses, including Qwen2-1.5B, Qwen2.5-7B, and Llama-3.1-8B. Details of the training and evaluation procedures for this section are provided in Appendix~\ref{apx:preliminary-implement}.

\subsection{Length Bias Indeed Exists}
\label{subsec:lenght-bias-in-model}

To verify the presence of length bias in reward models trained on $\mathcal{D}_{rm} = \{(x^{(i)}, y_w^{(i)}, y_l^{(i)})\}$, where $x^{(i)}$ is the prompt and response $y_w^{(i)}$ is preferred to $y_l^{(i)}$ according to human annotations, we design two additional evaluation settings based on $\mathcal{D}_{eval}$:
\begin{enumerate}
    \item Each prompt $x^{(i)}$ within $\mathcal{D}_{eval}$ is replaced with an empty prompt $x_e =$ {\em "empty prompt"}, forming the empty evaluation dataset $\mathcal{D}_{eval}^e$;
    \item Each prompt $x^{(i)}$ is randomly replaced with another prompt $x^{(j)}$ (where $i \neq j$)\footnote{For clarity of notation, we omit the superscript $^{(i)}$ in subsequent discussions where no ambiguity exists.}, resulting in the random evaluation dataset $\mathcal{D}_{eval}^r$.
\end{enumerate}

Table~\ref{tab:baseline-rm-original-empty-prompt-result} reports the evaluation results. Despite the semantic mismatch between responses and prompts in $\mathcal{D}_{eval}^e$ and $\mathcal{D}_{eval}^r$, the reward model nevertheless attains relatively high accuracy on these datasets. Specifically, nearly all models achieve an accuracy exceeding $60\%$, which is very close to the accuracy observed on the original dataset $\mathcal{D}_{eval}$. This finding suggests that there is a significant bias in the models towards preferring $y_w$, even when both responses are misaligned with the prompt.

Next, we examine the consistency of evaluation results across different reward models between $\mathcal{D}_{eval}^e$ (or $\mathcal{D}_{eval}^r$) and $\mathcal{D}_{eval}$, with a particular focus on the influence of varying prompts on the model's preferences. The results, as shown  in Table \ref{tab:baseline-rm-original-empty-prompt-result}, reveal that the models exhibit the same preference for responses in over $85\%$ of cases, despite different prompts. This suggests that the models' preferences are not primarily driven by the prompts.

\begin{figure*}[!ht]
    \centering
    \vspace{-0.2cm}
    \subfigure[Original evaluation dataset $\mathcal{D}_{eval}$]{\includegraphics[width=0.32\textwidth]{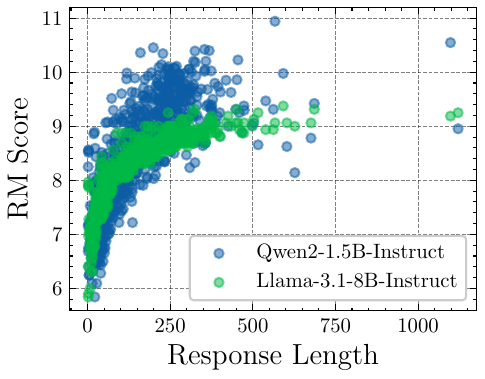}}
    \subfigure[Empty evaluation dataset $\mathcal{D}_{eval}^e$]{\includegraphics[width=0.32\textwidth]{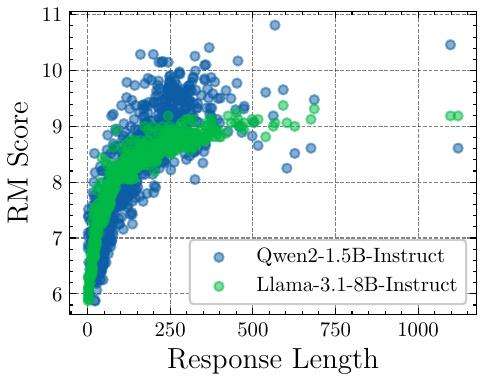}}
    \subfigure[Random evaluation dataset $\mathcal{D}_{eval}^r$]{\includegraphics[width=0.32\textwidth]{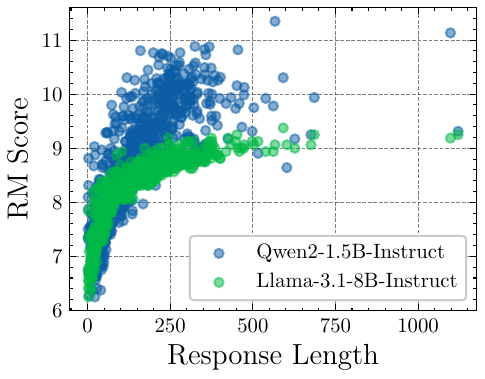}}
    \vspace{-0.2cm}
    \caption{The relationships between response lengths and scores of reward models (Baseline) trained with Qwen2-1.5B-Instruct and Llama-3.1-8B-Instruct, evaluated on different evaluation datasets, reveal a severe length bias.}
    \vspace{-0.4cm}
    \label{fig:length-score-ranges-baseline}
\end{figure*}

We further plot the relationship between response length and reward score for the reward model across $\mathcal{D}_{eval}$, $\mathcal{D}_{eval}^e$, and $\mathcal{D}_{eval}^r$, as shown in Figure \ref{fig:length-score-ranges-baseline}. A strong linear correlation is observed, suggesting that the reward model heavily relies on response length as a criterion for evaluating response quality. These findings provide compelling evidence of length bias in the reward models. More detailed analysis can be found in Appendix~\ref{appendix:lenght-bias-in-model}.

\subsection{Length Bias in Evaluation Dataset}
\label{subsec:length-bias-in-data}

Here, we aim to verify whether the initial evaluation dataset $\mathcal{D}_{eval}$ exhibits length bias, which could hinder its ability to accurately evaluate the true performance of the reward models. Specifically, our analysis reveals that $59.78\%$ of the chosen responses in $\mathcal{D}_{eval}$ are longer than the corresponding rejected responses. Consequently, this length bias allows the model to attain an accuracy nearing $60\%$ merely by favoring longer responses, resulting in an unreliable evaluation of the model's performance. 

To fairly and accurately assess the semantic quality of the reward model, we reconstruct the quality evaluation dataset $\mathcal{D}_{eval}^q$ from $\mathcal{D}_{eval}$ through an automated process. Specifically, we employ GPT-4o~\citep{hurst2024gpt} to transform each triplet $(x, y_w, y_l)$ in $\mathcal{D}_{eval}$ into two distinct triplets $(x, y_w^{1}, y_l^{1})$ and $(x, y_w^{2}, y_l^{2})$. These newly generated triplets adhere to the length constraints $|y_l^{1}| > |y_w^{1}|$\footnote{$|y|$ denotes the length of a given response $y$.} and $|y_l^{2}| < |y_w^{2}|$, while ensuring that $y_w^1$ and $y_w^2$, as well as $y_l^1$ and $y_l^2$, remain semantically similar to the original $y_w$ and $y_l$. The detailed generation process is documented in Appendix \ref{apx:construct-D_eval^q}. 

The reconstructed evaluation dataset $\mathcal{D}_{eval}^q$ demonstrates significant improvement in mitigating the inherent length bias in $\mathcal{D}_{eval}$,  thereby facilitating a more objective comparison among diverse reward models. As evidenced in Table~\ref{tab:rm-result},  under this less biased evaluation dataset, Baseline models exhibit notably reduced performance metrics, with the majority of accuracy scores not exceeding $60\%$. Further analysis of $\mathcal{D}_{eval}^q$ is in Appendix~\ref{appendix:length-bias-in-data}.

\subsection{Length Biased Reward Model Show Limited Adherence to Length Instructions}
\label{subsec:limit-length-following}

The reward model demonstrates the length bias, indicating its ability to perceive response length. However, the question remains whether it can apply this ability to follow length instruction $x_l$, where $x_l = \text{\em length constraint} + x$. For simplicity, we use the same {\em length constraint} as LIFT~\citep{yuan2024following}, which specified by {\em word\_num}, as detailed in Appendix \ref{apx:LIFT-decomposition}.
To validate this, we construct the length evaluation dataset $\mathcal{D}_{eval}^l$ based on $\mathcal{D}_{eval}^q$, where both responses semantically satisfy $x$ in $x_l$, but only one adheres to the length constraint. Appendix \ref{apx:construct-D_eval^l} for more construction details.

When evaluated on $\mathcal{D}_{eval}^l$ (results shown in Table \ref{tab:rm-result-length}, Baseline), all models demonstrate the accuracy close to $50\%$, no better than chance.
This demonstrates that models with length bias are unable to apply this ability to follow length instructions.

This finding suggests that reward models unconsciously acquire length bias during preference learning without explicit awareness of length as a measurable attribute. We hypothesize: \textbf{Could explicit length instruction learning enable models to develop clear perception of response length, thereby mitigating undesired length bias?} Based on this, we propose incorporating length constraints into the training process to transform implicit length bias into explicit length understanding.

\subsection{Length Instructions Are Easily Learned}
\label{subsec:length-following-ease-learn}

The most straightforward approach to explicit length instruction learning would be training directly on data formatted as $\{x_l, y_w, y_l\}$. To examine the effectiveness of this intuitive format, we extend LIFT~\citep{yuan2024following} to create LIFT-plus by incorporating minimum length constraints ("or more" length instruction) alongside its original maximum length constraints ("or less" length instruction). We then construct three variant datasets: 1) $\text{LIFT-plus}_2^{reverse}$, which reverses the preference order between chosen and rejected responses based on length constraints; 2) $\text{LIFT-plus}_2^{noreverse}$, which preserves the original preference ordering; and 3) $\text{LIFT-plus}_2^{empty}$, where substitutes $x$ in $x_l$ with an empty prompt $x_e$ while maintaining the preference order, examines the impact of semantic prompts on length instruction following. The complete construction process is detailed in Appendix~\ref{apx:LIFT-decomposition}.

Empirical results presented in Table~\ref{rm-length-prompt-result-reverse/unreverse/empty_prompt}~(Appendix \ref{appendix:length-following-ease-learn}) reveal that while models demonstrate notably higher accuracy on $\mathcal{D}_{eval}^l$,  they achieve diminished accuracy on $\mathcal{D}_{eval}^q$ compared to the Baseline in Table \ref{tab:rm-result}. This performance disparity suggests that reward models prioritize length instruction compliance at the expense of semantic understanding. Furthermore, our analysis of accuracy trajectories across training steps (Figure \ref{fig:length-quality-acc-train-loss-train-step}, Appendix \ref{appendix:length-following-ease-learn}) indicates a similar pattern: accuracy on $\mathcal{D}_{eval}^l$ exhibits rapid improvement, while accuracy on $\mathcal{D}_{eval}^q$ shows initial improvement followed by deterioration.

These results highlight the limitations of the $\{x_l, y_w, y_l\}$ format, which encourages overfitting to length instructions at the expense of semantic capability, indicating the need for better methods.
\section{Response-Conditioned Modeling}
\label{sec:method}

\begin{figure*}[!ht]
    \centering
    \vspace{-0.2cm}
    \includegraphics[width=0.9\textwidth]{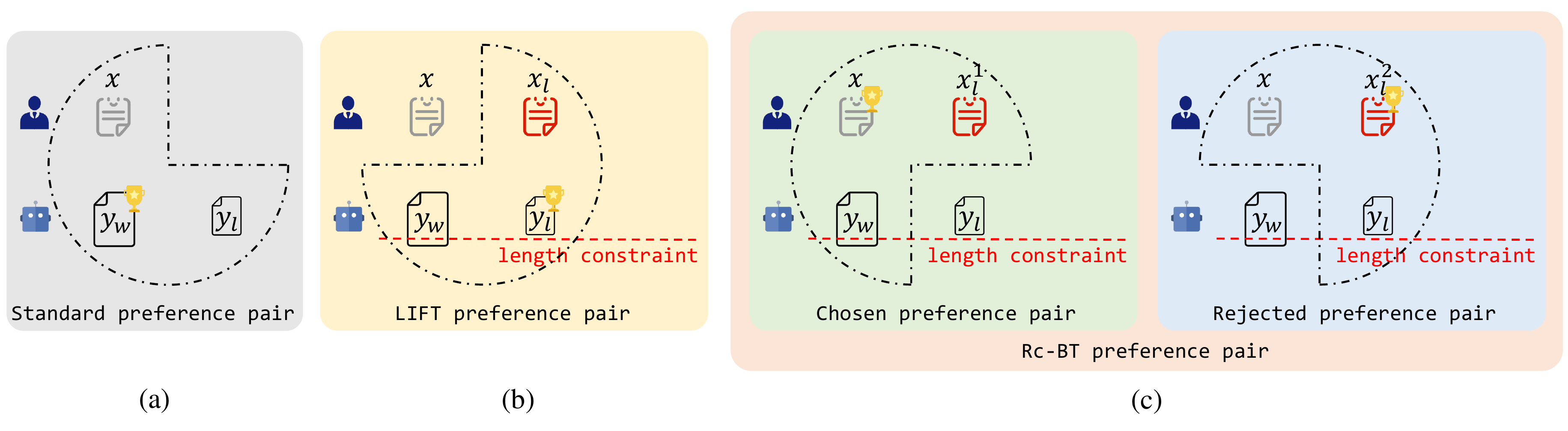}
    \vspace{-0.2cm}
    \caption{To illustrate the distinct data formats across different methods, we present the maximum length instruction case: \textbf{(a)} The conventional RLHF~\citep{ouyang2022training} with standard preference pair ($x$, $y_w$, $y_l$) ; \textbf{(b)} LIFT with augmented format  ($x_l$, $y_l$, $y_w$); and \textbf{(c)} Our method (\ourbtmodel{}) with two preference pairs ($x$, $x_l^1$, $y_w$) and ($x_l^2$, $x$, $y_l$). The term \textcolor{red}{length constraint} refers to the maximum allowable length of the response as specified in the length instruction $x_l$, $x_l^1$, or $x_l^2$. The black dashed lines indicate the data utilized by each method.}
    \label{fig:Rc-BT-overview}
   \vspace{-0.4cm}
\end{figure*}

In this section, we present our \ourbtmodel{}, with the overall framework depicted in Figure \ref{fig:Rc-BT-overview}. In Section~\ref{subsec:traditional-bt-model}, we briefly review the Bradley-Terry~(BT) model. In Section~\ref{subsec:Rc-BT}, we introduce the \textbf{R}esponse-\textbf{c}onditioned \textbf{B}radley-\textbf{T}erry~(\ourbtmodel{}) model\footnote{While primarily developed for length bias mitigation and length instruction adherence, the \ourbtmodel{} framework is theoretically applicable to diverse instruction-following tasks.}, presenting its algorithmic details and mathematical formulation. Finally, we demonstrate the implementation in both RM~(Section~\ref{subsec:Rc-RM}) and DPO~(Section~\ref{subsec:Rc-DPO}) through mathematical derivations.

\subsection{Preliminary: Bradley-Terry Model}
\label{subsec:traditional-bt-model}

In RLHF, given a human request $x$, a supervised fine-tuned LLM denoted as $\pi^{\text{SFT}}$ is prompted with $x$ to sample response pairs $y_1$, $y_2$. Human feedback on this pair is then collected and represented as $y_w \succ y_l$, where $y_w$ is preferred than $y_l$. To modeling human response preferences, the Bradley-Terry (BT)~\citep{bradley1952rank} model is a prevalent choice~\citep{ouyang2022training}. It models the underlining human preference distribution $p^*$ by assuming a underlining true reward  model $r^*$:

\begin{equation}
    p^*(y_w\succ y_l) = \frac{\exp{(r^*(x, y_w))}}{\exp(r^*(x, y_w))+\exp(r^*(x, y_l))},
    \label{eqn:response_dis}
\end{equation}

where $r^*$ is not accessible or easily defined by several rules. Therefore, people try to parameterize a reward model $r_\phi$ and optimize the parameters via maximum likelihood or leverage an analytical mapping from reward function to optimal policy, to directly optimize LLM policy~\citep{dubey2024llama}. Given a high quality preference dataset $\{(x, y_{w}, y_l)\}$, the BT model has demonstrated significant efficacy in various preference modeling and policy optimization tasks~\citep{NEURIPS2023_4dbb61cb}.

\subsection{Response-Conditioned BT Model}
\label{subsec:Rc-BT}

Response preference modeling employs Eqn.~\ref{eqn:response_dis} as a framework for response differentiation. However, a fundamental challenge arises in scenarios involving prompts with multiple instructions, where responses may only partially satisfy the given criteria. As previously discussed in Section \ref{subsec:length-following-ease-learn}, the conventional $\{x_l, y_w, y_l\}$ data format exhibits a tendency to overfit to length instructions, consequently compromising the model's semantic comprehension capabilities. This limitation represents an inherent constraint in response preference modeling.

Therefore, we propose a {\em response-conditioned modeling} framework, as illustrated in Figure \ref{fig:Rc-BT-overview}.  Our approach proceeds as follows: given an original chosen response $y_w$ and prompt $x$, we construct a {length augmented instruction} $x_l^1$ by incorporating a {\em length constraint} with the prompt $x$, ensuring that $y_w$ violates this constraint. We then formulate a preference pair $(x, x_l^1, y_w)$, where $(x, y_w)$  is considered preferable to $(x_l^1, y_w)$. This preference is then modeled using the BT formulation as follow:

{\small
    \begin{equation}
        p^*(x\succ x_l^1|y_w) = \frac{\exp{(r^*(x, y_w))}}{\exp(r^*(x, y_w))+\exp(r^*(x_l^1, y_w))}.
        \label{eqn:prompt_dis_chosen}
    \end{equation}
}

Similarly, for a given rejected response $y_l$ and original prompt $x$, we construct a length augmented instruction $x_l^2$ such that $y_l$ satisfies the specified length constraint. We then formulate a preference pair $(x_l^2, x, y_l)$, where $(x_l^2, y_l)$ is considered preferable to $(x, y_l)$. This preference structure is similarly modeled using the BT formulation:

{\small
    \begin{equation}
        p^*(x_l^2\succ x|y_l) = \frac{\exp{(r^*(x_l^2, y_l))}}{\exp(r^*(x_l^2, y_l))+\exp(r^*(x, y_l))}.
        \label{eqn:prompt_dis_reject}
    \end{equation}
}

The final response-conditioned modeling preference dataset is defined as $\mathcal{D}_{Rc} = \{(x, x_l^1, y_w)\} \cup \{(x_l^2, x, y_l)\}$. Through maximum likelihood estimation, we derive the Response-conditioned BT (\ourbtmodel{}) modeling objective function as follows:

\begin{equation}
    \begin{aligned}
        \mathcal{L}_{Rc} = & -\underset{(x, x_l^1, y_w) \sim \mathcal{D}_{Rc}}{\mathbb{E}}[\log p^*(x\succ x_l^1|y_w)] \\& - \underset{(x_l^2, x, y_l) \sim \mathcal{D}_{Rc}}{\mathbb{E}}[\log p^*(x_l^2\succ x|y_l)].
    \end{aligned}    
    \label{eqn:prompt_dis}
\end{equation}

The \ourbtmodel{} facilitates explicit comparisons between $x$ and its length-augmented variants ($x_l^1$ and $x_l^2$), enabling the model to systematically perceive response lengths and thus mitigating implicit length bias through explicit length understanding.

\subsection{Response-Conditioned Reward Model}
\label{subsec:Rc-RM}

In response preference modeling, the reward model is initialized from $\pi^\text{SFT}$ and augmented with a linear projection layer that transforms the complete sequence representation into a scalar value $r_\phi(x, y)$. Given a preference dataset $\mathcal{D}=\{(x, y_{w}, y_l)\}$, the reward model is optimized through maximum likelihood estimation according to Eqn.~\ref{eqn:response_dis}:

{\small
    \begin{equation}
        \mathcal{L}_{r_\phi}(\mathcal{D}) = -\underset{(x, y_{w}, y_l) \sim \mathcal{D}}{\mathbb{E}}[ \log\sigma (r_\phi(x, y_{w})- r_\phi(x, y_l))],
        \label{eqn:prompt-conditioned-rm}
    \end{equation}
}

\noindent
where $\sigma(\cdot)$ is the sigmoid function.

In response-conditioned modeling, the architectural structure of the parameterized reward model $r_\phi$ remains unchanged while the data format is transformed from prompt-conditioned to response-conditioned. Similar to Eqn.~\ref{eqn:prompt-conditioned-rm}, the model is optimized through maximum likelihood estimation based on Eqn.~\ref{eqn:prompt_dis} using dataset $\mathcal{D}_{Rc}$:

{\small
    \begin{equation}
        \begin{aligned}
            \mathcal{L}_{r_\phi}(\mathcal{D}_{Rc}) = & -\underset{(x, x_l^1, y_w) \sim \mathcal{D}_{Rc}}{\mathbb{E}}[\log\sigma (r_\phi(x, y_w) - r_\phi(x_l^1, y_w))] \\ & - \lambda \underset{(x_l^2, x, y_l) \sim \mathcal{D}_{Rc}}{\mathbb{E}}[\log\sigma (r_\phi(x_l^2, y_l) - r_\phi(x, y_l))],
        \end{aligned}
        \label{eqn:response-conditioned-rm}
    \end{equation}
}

\noindent
where $\lambda$ is used to balance the relative contribution of the pairs $(x, x_l^1, y_w)$ and $(x_l^2, x, y_l)$.

\subsection{Response-Conditioned Direct Preference Optimization}
\label{subsec:Rc-DPO}
In response preference modeling, DPO derives an alternative formulation from Eqn.~\ref{eqn:prompt-conditioned-RL}, where the reward is expressed as a function of the optimal policy. By substituting this formulation into the reward optimization objective specified in Eqn. \ref{eqn:prompt-conditioned-rm}, we obtain a direct optimization approach for policy training. Specifically, the policy can be optimized on dataset $\mathcal{D}$ using the following objective function:

{\small
    \begin{equation}
        \begin{aligned}
           & \mathcal{L}_{DPO}(\pi_\theta; \pi_\text{ref}) = \\ & -\underset{(x, y_w, y_l) \sim \mathcal{D}}{\mathbb{E}}[\log\sigma(\beta \log\frac{\pi_\theta(y_w|x)}{\pi_\text{ref}(y_w|x)} - \beta \log\frac{\pi_\theta(y_l|x)}{\pi_\text{ref}(y_l|x)})].
        \end{aligned}
         \label{eqn:origin-dpo}
    \end{equation}
}

For the response-conditioned modeling, we follow an analogous derivation process to DPO. Specifically, we begin with the modified RL objective (Eqn.~\ref{eqn:response-conditioned-RL}), derive the reward model expression in terms of the optimal policy, and subsequently incorporate it into the \ourbtmodel{} modeling function (Eqn.~\ref{eqn:prompt_dis}). This derivation yields the Response-conditioned DPO (Rc-DPO) objective function:

{\small
    \begin{equation}
        \begin{aligned}
            & \mathcal{L}_{DPO}^{Rc}(\pi_\theta; \pi_\text{ref}) = \\ & -\underset{(x, x_l^1, y_w) \sim \mathcal{D}_{Rc}}{\mathbb{E}} [\log\sigma(\beta \log\frac{\pi_\theta(x, y_w)}{\pi_\text{ref}(x, y_w)} - \beta \log\frac{\pi_\theta(x_l^1, y_w)}{\pi_\text{ref}((x_l^1, y_w)})] \\ & -\underset{((x_l^2, x, y_l) \sim \mathcal{D}_{Rc}}{\mathbb{E}} [\log\sigma(\beta \log\frac{\pi_\theta((x_l^2, y_l)}{\pi_\text{ref}(x_l^2, y_l)}) - \beta \log\frac{\pi_\theta(x, y_l)}{\pi_\text{ref}(x, y_l)}].
        \end{aligned}
        \label{eqn:response-conditioned-dpo}
    \end{equation}
}
\noindent
See Appendix~\ref{apx:rc-bt-derivation} for a complete derivation. 
\section{Experiments}
\label{sec:experiments}

This section presents experiments and ablation studies to demonstrate the effectiveness and validate the design of our \ourbtmodel{} framework. We primarily demonstrate the results of our method in mitigating length bias, while its efficacy in length instruction following can be found in Appendix~\ref{apx:additional-experimental-results}.

\subsection{Experimental Settings}
\label{subsec:experiment-settings}
{\bf Dataset and Models.} Consistent with Section~\ref{sec:preliminary}, for the dataset, we use $\mathcal{D}_{sft}$ for SFT to finetune the {\em Base} models. For RM and DPO, \ourbtmodel{} generates an augmented dataset $\mathcal{D}_{Rc}$ derived from $\mathcal{D}_{rm}$. Both {reward models} and DPO models are trained on the combined dataset $\mathcal{D}_{rm} \cup \mathcal{D}_{Rc}$, referred to as Rc-RM and Rc-DPO, respectively.
We employ three pretrained models as our base models: Qwen2-1.5B\footnote{We also conduct experiments using the Qwen2.5-1.5B models, which shown in Appendix~\ref{qwen2.5-1.5b-results}.}, Qwen2.5-7B, and Llama-3.1-8B.

{\bf Training Details.} For Rc-RM training, the learning rate is set to $1 \times 10^{-5}$, followed by a cosine learning rate schedule with an initial warmup of $10$ steps and a batch size of $64$. Each experiment is trained for $5$ epochs. For Rc-DPO training, the settings are identical to RM training, except the learning rate is $1 \times 10^{-6}$. All experiments are implemented based on DeepSpeed~\citep{yao2023deepspeed} and Transformers~\citep{wolf2020transformers}, and conducted on a machine with $8$ NVIDIA A$100$ $80$GB GPUs. 

\begin{table*}[!ht]
    \centering
    \small
    \vspace{-0.3cm}
    \caption{Evaluation results of different DPO models on AlpacaEval \citep{dubois2024alpacafarm}.}
    \begin{tabular}{ccccccccc}
        \toprule
        \multirow{2}{*}{Metrics} & \multicolumn{4}{c}{Qwen2.5-7B-Base} & \multicolumn{4}{c}{Qwen2.5-7B-Instruct} \\
        \cmidrule(lr){2-5} \cmidrule(lr){6-9} & Baseline & LIFT-plus & R-DPO & Rc-DPO & Baseline & LIFT-plus & R-DPO & Rc-DPO \\
        \cmidrule(lr){1-1} Quality Win Ratio (\%) & 33.54 & 31.67 & 40.40 & \textbf{45.39} & 28.43 & 25.69 & 34.16 & \textbf{44.63} \\
        Response Length & 517.30 & 184.17 & 583.18 & 208.42 & 261.32 & 195.80 & 235.39 & 228.24 \\ 
        \midrule
        \multirow{2}{*}{Metrics} & \multicolumn{4}{c}{Llama-3.1-8B-Base} & \multicolumn{4}{c}{Llama-3.1-8B-Instruct} \\
        \cmidrule(lr){2-5} \cmidrule(lr){6-9} & Baseline & LIFT-plus & R-DPO & Rc-DPO & Baseline & LIFT-plus & R-DPO & Rc-DPO \\
        \cmidrule(lr){1-1} Quality Win Ratio (\%) & 46.30 & 40.15 & 49.13 & \textbf{58.10} & 42.52 & 47.88 & 42.64 & \textbf{64.34} \\
        Response Length & 435.98 & 157.80 & 465.72 & 202.01 & 247.74 & 153.77 & 215.82 & 204.77 \\ 
        \bottomrule
    \end{tabular}
    \vspace{-0.3cm}
    \label{tab:dpo-origin-quality-result}
\end{table*}

{\bf Compared Methods and Evaluation.}
For RM evaluation, we assess the performance across methods using two primary metrics: {\em Quality Eval Acc} and {\em Length Eval Acc}, which are accuracy on $\mathcal{D}_{eval}^q$ and $\mathcal{D}_{eval}^l$ respectively. The former metric evaluates semantic quality with a focus on length bias mitigation, while the latter measures the effectiveness of length instruction adherence.

For DPO evaluation, leveraging recent advancements in automated assessment approaches~\citep{zheng2023judging, dubois2024alpacafarm}, we employ a model-based evaluation framework to systematically assess the quality of generated responses. First, we construct the AlpacaEval-LI-plus-less and AlpacaEval-LI-plus-more benchmarks, enhanced versions of AlpacaEval~\citep{dubois2024alpacafarm}, designed to evaluate the model's ability to follow "or less" and "or more" length instructions, respectively. Then, we evaluate each DPO model's performance by comparing it with its corresponding SFT/{\em Instruct} model using two complementary metrics. The first metric, {\em Length Acc}, quantifies the model's adherence to length instructions by measuring how well the generated responses conform to the specified length constraints in $x_l$. The second metric assesses the model's semantic instruction following capability through two comparative win ratios: {\em Length Win Ratio} represents the proportion of cases where the model generates semantically superior responses when prompted with length-augmented instructions ($x_l$), while {\em Quality Win Ratio} measures the proportion of cases where the model demonstrates better semantic quality with original prompts ($x$). For comprehensive evaluation details, please refer to Appendix \ref{apx:dpo-eval-details}.

We refer to the model trained on the $\mathcal{D}_{rm}$ as {\bf Baseline}. For RM, we incorporate {\bf ODIN}~\citep{chenodin} as comparison method, with particular emphasis on its {\em Quality Eval Acc}, owing to its demonstrated effectiveness in mitigating length bias. In addition, we compare our method with {\bf LIFT-plus} on the {\em Length Eval Acc}\footnote{As discussed in Section~\ref{sec:preliminary}, LIFT-plus harms semantic quality and does not mitigate length bias, so we exclude it from the \QEA{} comparison.}. For DPO, we include {\bf LIFT-plus} and {\bf R-DPO}~\citep{park2024disentangling} as comparison methods. Unless otherwise specified, all compared methods are trained using the recommended settings from their papers.

\subsection{The Results of Reward Models}
\label{subsec:rm-results}

\begin{table}[H]
    \centering
    \small
    \vspace{-0.3cm}
    \caption{\QEA{} of different reward models on quality ($\mathcal{D}_{eval}^q$) evaluation datasets.}
    \vspace{-0.2cm}
    \label{tab:rm-result}
    \begin{tabular}{lccc}
            \toprule
            Model & Baseline & ODIN & Rc-RM \\
            \midrule
            Qwen2-1.5B-Base & 59.14 & 56.12 & \textbf{69.55} \\
            Qwen2-1.5B-Instruct & 60.75 & 63.56 & \textbf{71.47} \\
            \midrule
            Qwen2.5-7B-Base & 54.26 & 60.90 & \textbf{70.74} \\
            Qwen2.5-7B-Instruct & 59.31 & 67.55 & \textbf{73.07} \\
            \midrule
            Llama-3.1-8B-Base & 59.04 & 60.37 & \textbf{70.51} \\
            Llama-3.1-8B-Instruct & 55.59 & 60.90 & \textbf{72.44} \\
            \bottomrule
        \end{tabular}
    \vspace{-0.2cm}
\end{table}

{\bf Rc-RM significantly alleviates length bias.} The {\em Quality Eval Acc} results in Table \ref{tab:rm-result} clearly demonstrate the effectiveness of \ourrm{} in mitigating length bias. Specifically, \ourrm{} outperforms both Baseline and ODIN across all models. Notably, on Qwen2-1.5B-Base, \ourrm{} exceeds Baseline by $10.41\%$ and outperforms ODIN by $13.43\%$, respectively, while on Llama-3.1-8B-Instruct, it surpasses Baseline by $16.85\%$ and ODIN by $11.54\%$.

\begin{table}[H]
    \centering
    \small
    \vspace{-0.2cm}
    \caption{Evaluation results of different PPO models on AlpacaEval~\citep{dubois2024alpacafarm}.}
    \vspace{-0.2cm}
    \begin{tabular}{cccc}
        \toprule
        \multirow{2}{*}{Metrics} & \multicolumn{3}{c}{Qwen2-1.5B-Instruct} \\
        \cmidrule(lr){2-4} & Baseline & ODIN & Rc-RM \\
        \cmidrule(lr){1-1} Quality Win Ratio (\%) & 38.28 & 41.57 & \textbf{44.06} \\
        Response Length & 671.49 & 342.85 & 308.16 \\ 
        \midrule
        \multirow{2}{*}{Metrics} & \multicolumn{3}{c}{Qwen2.5-7B-Instruct} \\
        \cmidrule(lr){2-4} & Baseline & ODIN & Rc-RM \\
        \cmidrule(lr){1-1} Quality Win Ratio (\%) & 40.31 & 45.03 & \textbf{49.26} \\
        Response Length & 597.27 & 336.15 & 277.15 \\ 
        \bottomrule
    \end{tabular}
    \vspace{-0.2cm}
    \label{tab:ppo-origin-quality-result}
\end{table}

We conducted further experiments by training a policy using the trained reward models via PPO. We trained the {\em Qwen2-1.5B-Instruct} policy using the reward models of Qwen2-1.5B-Instruct and Qwen2.5-7B-Instruct from Table~\ref{tab:rm-result}. The evaluation was carried out on AlpacaEval following the same metrics used for evaluating the DPO models. The results are in Table~\ref{tab:ppo-origin-quality-result}. It can be observed that the policy trained using Rc-RM exhibits high semantic quality. Its performance significantly surpasses other length-bias-mitigating methods, further demonstrating the effectiveness of our method.

\begin{figure}[!ht]
    \centering
    \vspace{-0.2cm}
    \subfigure[Qwen2-1.5B-Instruct]{\includegraphics[width=0.232\textwidth]{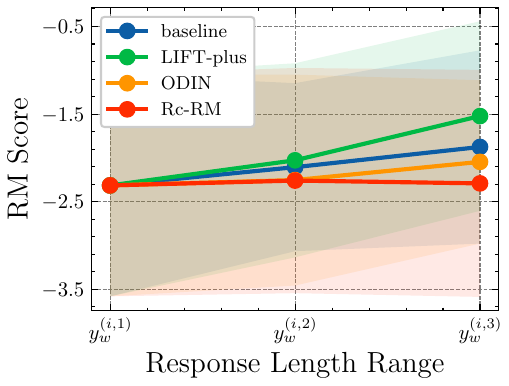}}
    \subfigure[Llama-3.1-8B-Instruct]{\includegraphics[width=0.228\textwidth]{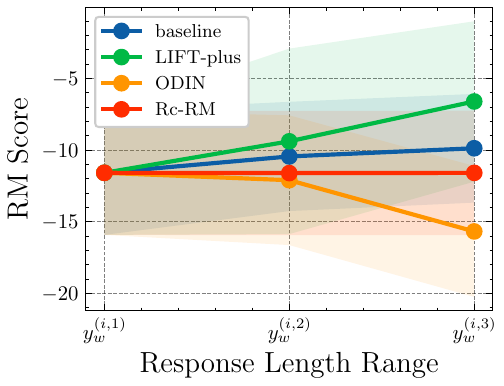}}
    \vspace{-0.2cm}
    \caption{Analysis of reward scores across  models on  $\mathcal{D}_{eval}^{ml}$ as a function of response length, with smaller changes indicating reduced length bias.}
    \label{fig:multi-length-bias-multi-wordnum-rm-score-diff}
    \vspace{-0.4cm}
\end{figure}

Furthermore, for $\mathcal{D}_{eval}$, we rewrite each $(x^{(i)},\allowbreak y_w^{(i)})$ into multiple length-varied pairs {$\mathcal{D}_{eval}^{ml} \allowbreak = \{(x^{(i)}, y_w^{(i, j)})\}, j\in\{1,2,3\}$}, where $|y_w^{(i, 1)}| < |y_w^{(i, 2)}| < |y_w^{(i, 3)}|$, while maintaining the semantic consistency between $y_w^{(i)}$ and $y_w^{(i, j)}$. Then we evaluate each {reward model} on $\mathcal{D}_{eval}^{ml}$, with the results shown in Figure~\ref{fig:multi-length-bias-multi-wordnum-rm-score-diff}, where the reward model with less length bias exhibits smaller slopes in its scores.
The results indicate that both Baseline and LIFT-plus show an upward trend in their scores, while ODIN fluctuates due to varying degrees of length penalty. \ourrm{}, however, demonstrates superior stability, clearly highlighting its effectiveness in mitigating length bias.

\subsection{The Results of DPO Models}
\label{subsec:dpo-results}

{\bf Rc-DPO effectively reduces response length and improves quality.} Table~\ref{tab:dpo-origin-quality-result} highlights the effectiveness of \ourdpo{} in mitigating length bias and enhancing response quality on AlpacaEval. \ourdpo{} significantly outperforms all comparison methods in semantic quality. Specifically, LIFT-plus negatively impacts semantic quality, leading to a decline in {\em Quality Win Ratio}. Compared to R-DPO, \ourdpo{} not only achieves a notable improvement in {\em Quality Win Ratio} but also reduces {\em Response Length}, demonstrating its ability to alleviate length bias and improve the semantic quality.

\subsection{Ablation Studies}
\label{ablations}
Due to the high cost of evaluating the DPO model, we focus on the reward model for ablation studies. For convenience, we denote $\{(x, x_l^1, y_w)\}$ as $\mathcal{D}_{Rc}^c$ and $\{(x_l^2, x, y_l)\}$ as $\mathcal{D}_{Rc}^r$. A more detailed analysis can be found in Appendix~\ref{appendix-ablation}.

{\bf $\mathcal{D}_{Rc}^c$ and $\mathcal{D}_{Rc}^r$ are complementary.} We conduct ablation experiments by training with $\mathcal{D}_{rm} \cup \mathcal{D}_{Rc}^c$ ({\em w/o} $\mathcal{D}_{Rc}^r$) and $\mathcal{D}_{rm} \cup \mathcal{D}_{Rc}^r$ ({\em w/o} $\mathcal{D}_{Rc}^c$), with the results shown in Table~\ref{tab:rm-result-ablation-Rc-BT}. As observed, when only $\mathcal{D}_{Rc}^c$ or $\mathcal{D}_{Rc}^r$ is used, the {\em Quality Eval Acc} of \ourrm{} drops significantly, nearing the performance of Baseline in Table~\ref{tab:rm-result}, while its {\em Length Eval Acc} hovers around $50\%$, failing to learn length instructions. Subsequently, we prepend a length constraint to each $x$ in $\mathcal{D}_{eval}^q$, forming $x_l$, where $y_w$ satisfies the length constraint, resulting in $\mathcal{D}_{eval}^{q,l}$. The evaluation results on both $\mathcal{D}_{eval}^q$ and $\mathcal{D}_{eval}^{q,l}$, shown in Figure~\ref{fig:llama-3.1-8b-instruct-origin-length-score-ranges-ablation}. Specifically, Figure~\ref{fig:llama-3.1-8b-instruct-origin-length-score-ranges-ablation}(a) indicates that for the reward models trained with $\mathcal{D}_{rm} \cup \mathcal{D}_{Rc}^{c}$, the scores of $(x_l, y_w)$ are consistently lower than those of the original $(x, y_w)$, despite $y_w$'s length satisfies {$x_l$}. A similar phenomenon is observed in $\mathcal{D}_{rm} \cup \mathcal{D}_{Rc}^{r}$ (Figure~\ref{fig:llama-3.1-8b-instruct-origin-length-score-ranges-ablation}(b)). Therefore, the combination of $\mathcal{D}_{Rc}^c$ and $\mathcal{D}_{Rc}^r$ is essential for preventing the reward model from developing new biases.

\begin{table}[!ht]
\small
    \centering
    \vspace{-0.2cm}
    \caption{Evaluation results of Rc-RM {\em w/o} $\mathcal{D}_{Rc}^c$ and {\em w/o} $\mathcal{D}_{Rc}^r$ on quality ($\mathcal{D}_{eval}^q$) and length ($\mathcal{D}_{eval}^l$) evaluation datasets.}
    \vspace{-0.2cm}
    \scalebox{0.8}{
    \begin{tabular}{ccccc}
        \toprule
        \multirow{2}{*}{Metrics} & \multicolumn{2}{c}{Qwen2-1.5B-Base} & \multicolumn{2}{c}{Qwen2-1.5B-Instruct} \\
        \cmidrule(lr){2-3} \cmidrule(lr){4-5} & {\em w/o} $\mathcal{D}_{Rc}^c$ & {\em w/o} $\mathcal{D}_{Rc}^r$ & {\em w/o} $\mathcal{D}_{Rc}^c$ & {\em w/o} $\mathcal{D}_{Rc}^r$ \\
        \cmidrule(lr){1-1} Quality Eval Acc ($\%$) & 58.78 & 55.59 & 59.31 & 57.71 \\
        Length Eval Acc ($\%$) & 45.83 & 36.22 & 45.51 & 44.87 \\ 
        \midrule
        \multirow{2}{*}{Metrics} & \multicolumn{2}{c}{Llama-3.1-8B-Base} & \multicolumn{2}{c}{Llama-3.1-8B-Instruct} \\
        \cmidrule(lr){2-3} \cmidrule(lr){4-5} & {\em w/o} $\mathcal{D}_{Rc}^c$ & {\em w/o} $\mathcal{D}_{Rc}^r$ & {\em w/o} $\mathcal{D}_{Rc}^c$ & {\em w/o} $\mathcal{D}_{Rc}^r$ \\
        \cmidrule(lr){1-1} Quality Eval Acc ($\%$) & 50.27 & 50.53 & 58.51 & 52.13 \\
        Length Eval Acc ($\%$) & 44.87 & 32.69 & 37.18 & 33.65 \\ 
        \bottomrule
    \end{tabular}
    }
    \vspace{-0.4cm}
    \label{tab:rm-result-ablation-Rc-BT}
\end{table}

\section{Conclusion}
\label{sec:conclusion}
We propose \ourbtmodel{}, a method that separates human semantic intent from length instructions to mitigate length bias while retaining length sensitivity, and provide mathematical extensions to RM and DPO. Extensive experiments demonstrate the effectiveness of \ourbtmodel{} in reducing length bias and improving length instruction following. 

\section*{Limitations}
\label{sec:limitations}

\textbf{Additional computational cost introduced by data augmentation.} Our framework creates a response-conditioned augmentation set $\mathcal{D}_{Rc}$ and trains on the union $\mathcal{D}_{rm} \cup \mathcal{D}_{Rc}$. This roughly doubles the number of preference pairs and requires training both reward and policy models. Although the augmented data demonstrate high sample efficiency, the consequent increase in computational resources required may present modest constraints for researchers with limited computational budgets. \textbf{Evaluation confined to smaller models.} All experiments are conducted on Qwen2-1.5B, Qwen2.5-1.5B, Qwen2.5-7B and Llama-3.1-8B models, leaving larger models (e.g., 13B, 34 B, 70B) unexplored. Extending the study to higher-capacity LLMs is an important avenue for future work.

\bibliography{custom}

\appendix

\section{Related Work}
\label{apx:full-related work}
{\bf Reinforcement Learning from Human Feedback.} Reinforcement Learning from Human Feedback (RLHF)~\citep{ziegler2019fine} has established itself as a dominant paradigm for aligning LLMs with human preferences~\citep{achiam2023gpt, team2023gemini, caruccio2024claude, yang2024qwen2, dubey2024llama}. At its core, RLHF trains models by optimizing reward signals derived from human feedback, typically collected through preference comparisons or explicit ratings~\citep{bai2022constitutional, lee2023rlaif}. This approach has demonstrated significant efficacy in refining LLM behavior to align with human expectations across diverse complex tasks~\citep{kreutzer-etal-2018-reliability, liu2020learning, ziegler2019fine, ouyang2022training}.

However, RLHF necessitates training a reward model (RM) to approximate human preferences, followed by LLM alignment through reinforcement learning (RL) algorithms, notably Proximal Policy Optimization (PPO)~\citep{schulman2017proximal}. This process presents significant challenges and stability issues~\citep{NEURIPS2021_63c4b1ba, 9501950}. Direct preference alignment methods, in contrast, provide a more robust training approach~\citep{ivison2024unpacking, hong2024orpo, xu2024contrastivepreferenceoptimizationpushing}. Among these methods, direct preference optimization (DPO)~\citep{rafailov2024direct} has emerged as a leading technique, inspiring various derivative algorithms~\citep{hong2024orpo, chen2024noise, ethayarajh2024kto}. By reformulating the RLHF reward function, DPO eliminates both the need for an online reward model and the instabilities inherent in RL algorithms, enabling stable offline preference learning.

Our method enhances RLHF by improving reward model robustness and seamlessly integrates into DPO, enabling more effective LLM training.

{\bf Reward Hacking.} Despite its promising performance, RLHF is vulnerable to reward hacking --- the over-optimization of the reward model~\citep{skalse2022defining, pan2022effects, casper2023open, lambert2023alignment}. This phenomenon occurs when the policy exploits inherent biases in the reward model to maximize rewards without achieving intended objectives. Similar exploitation patterns have been observed in DPO-trained LLMs~\citep{lambert2024rewardbench, xu2024is}. The root causes are multifaceted: task complexity and subjectivity~\citep{casper2023open}, evaluation criteria limitations~\citep{parrish-etal-2022-bbq}, and evaluator qualification constraints~\citep{skalse2022defining}. Consequently, human preference data exhibit biases and inconsistencies~\citep{dubois2024alpacafarm}, and reward models struggle with preference approximation and out-of-distribution (OOD) generalization.

One of the most common forms of reward hacking is length bias \citep{shen2023loose, singhal2023long, park2024disentangling, chenodin}, where the reward models tend to favor longer responses, assigning higher reward scores to longer responses even if their semantic contents are not of high quality. This tendency is largely driven by human evaluators' preference for longer answers, which is easily exploited by the LLMs. As a result, LLMs often generate unnecessarily lengthy replies to appear more detailed or better formatted, even when the actual quality remains unchanged. To address this issue, several methods have been proposed. ODIN \citep{chenodin} adds multiple length regularization terms to the RM's training objective, which helps to distinguish response quality from length and effectively controls the LLM's dependency on response length. \citet{shen2023loose} involve using dual-reward model and different sets of learning rate hyperparameters, allowing different part of the reward model to learn distinct paradigms, thereby removing the length bias from final reward scores. Length Regularized DPO (R-DPO) \citep{park2024disentangling} modify the DPO training objective by adding a length penalty term to prevent length bias exploitation. 

However, the above methods assume that the model's sensitivity to length information is detrimental and, therefore, seek to eliminate the model's perception of length information. In contrast, our approach enables the model to explicitly distinguish between human semantic intent and length instruction, preserving its ability to perceive response length rather than completely eliminating the perception of length information. 

{\bf Length Instruction Following.} Current state-of-the-art (SOTA) LLMs, both open-source and closed-source, demonstrate a certain degree of implicit ability to follow length instructions \citep{yuan2024following}. For example, adding terms like "concise" or "verbose" can influence the length of the model's outputs. However, when it comes to explicit length instructions, such as "Answer the following instruction using $150$ words or less.", these models often fail to adhere effectively. To address this, LIFT \citep{yuan2024following} enhances the model's length instruction following capability by constructing an explicit length instruction preference dataset. However, the dataset used by LIFT is built at the expense of semantic quality, resulting in a degradation of the model's output quality. In contrast, our model not only effectively learns length instructions but also maintains, even in many cases improves, the semantic quality of the model's outputs. 

\clearpage\onecolumn
\section{Derivation of the DPO Objective Under the Response-Conditioned Bradley-Terry Model}
\label{apx:rc-bt-derivation}
{\bf Derivation of RL Fine-Tuning Objective.} During the RL phase of traditional RLHF, the trained reward model $r_\phi$ serves as the feedback mechanism for policy optimization. The standard RL optimization objective is formulated as:

\begin{equation}
    \max_{\pi_{\theta}} \mathbb{E}_{x\sim \mathcal{D}, y\sim \pi_{\theta}(y \mid x)}\bigl[r_{\phi}(x, y)\bigr] - \beta\mathbb{D}_{\textrm{KL}}\bigl[\pi_{\theta}(y\mid x)\mid \mid \pi_{\text{ref}}(y\mid x)\bigr],
    \label{eqn:prompt-conditioned-RL}
\end{equation}

under the reference policy $\pi_\text{ref}$ and the parametrized policy $\pi_\theta$. In the response-conditioned scenario, where responses are predetermined and guide prompt selection, the KL-divergence constraint in Eqn. \ref{eqn:prompt-conditioned-RL} requires modification. Accordingly, we reformulate the response-conditioned RL optimization objective as:

\begin{equation}
    \max_{\pi_{\theta}} \mathbb{E}_{y\sim \mathcal{D}_{Rc}, x\sim \pi_{\theta}(x \mid y)}\bigl[r_{\phi}(x, y)\bigr] - \beta\mathbb{D}_{\textrm{KL}}\bigl[\pi_{\theta}(x\mid y)\mid \mid \pi_{\text{ref}}(x\mid y)\bigr].
    \label{eqn:response-conditioned-RL}
\end{equation}

For simplicity, we unify the notation of $\mathcal{D}_{Rc} = \{(x, x_l^1, y_w)\} \cup \{(x_l^2, x, y_l)\}$ as $\{(x_w, x_l, y)\}$. Next, similar to the derivation in DPO \citep{park2024disentangling}, we can derive the partition function $Z(y)$ from Eqn. \ref{eqn:response-conditioned-RL}:

\begin{equation}
    Z(y) =\sum_{x}\pi_{\text{ref}}(x\mid y)\exp\left(\frac{1}{\beta}r_\phi(x, y)\right).
    \label{eqn:partition-function}
\end{equation}

Since the partition function is a function only of the response $y$ and the reference policy $\pi_{\text{ref}}$, and does not depend on the prompt $x$ or the parametrized policy $\pi_\theta$ to be optimized, we can reorganize Eqn. \ref{eqn:response-conditioned-RL} to obtain the following objective:

\begin{align}
    \min_{\pi_\theta}\mathbb{E}_{y\sim \mathcal{D}}\left[\mathbb{E}_{x\sim \pi_\theta(x|y)}\left[\log\frac{\pi_\theta(x|y)}{\pi^*(x|y)}\right] - \log Z(y)\right]=\min_{\pi_\theta}\mathbb{E}_{y\sim\mathcal{D}}\left[\mathbb{D}_{\text{KL}}(\pi_\theta(x|y)\mid\mid\pi^*(x|y)) - \log Z(y)\right],
    \label{eqn:proxy_policy}
\end{align}

\begin{equation}
    \pi^*(x\mid y) = \frac{1}{Z(y)}\pi_{\text{ref}}(x\mid y)\exp\left(\frac{1}{\beta}r_\phi(x, y)\right),
\end{equation}

where $\pi^*(x\mid y)$ is a valid probability distribution which $\pi^*(x\mid y) > 0$ and $\sum_x\pi^*(x\mid y) = 1$. By using Gibbs' inequality, the KL divergence is minimized to $0$ if and only if the two distributions are identical. Therefore, we obtain the optimal solution:

\begin{equation}
    \pi_\theta(x\mid y) = \pi^*(x\mid y) = \frac{1}{Z(y)}\pi_{\text{ref}}(x\mid y)\exp\left(\frac{1}{\beta}r_\phi(x, y)\right).
    \label{eqn:op_policy}
\end{equation}

Finally, by taking the logarithm on both sides and performing some basic algebraic operations, we can express the reward model $r_\phi(x, y)$ through its corresponding optimal policy $\pi^*(x\mid y)$:

\begin{equation}
    r_\phi(x,y) =\beta \log \frac{\pi^*(x\mid y)}{\pi_{\text{ref}}(x\mid y)} + \beta \log Z(y).
    \label{eqn:main_eq}
\end{equation}

{\bf Derivation of Response-conditioned DPO Objective.} Following Section \ref{subsec:Rc-BT}, the Response-conditioned Bradley-Terry model is formulated as:

\begin{equation}
    p^*(x_w\succ x_l|y) = \frac{\exp{(r^*(x_w, y))}}{\exp(r^*(x_w, y))+\exp(r^*(x_l, y))}.
    \label{eqn:prompt_dis_chosen_repeat}
\end{equation}

In the above derivation, we demonstrated that the optimal reward $r^*(x, y)$, parameterized as $r_\phi(x, y)$, can be expressed in term of its corresponding optimal policy $\pi^*(x\mid y)$. By substituting Eqn.~\ref{eqn:main_eq} into Eqn.~\ref{eqn:prompt_dis_chosen_repeat} and preforming simple simplifications, we obtain:

\begin{align}
    p^*(x_w\succ x_l|y)&=\frac{\exp\left(\beta \log \frac{\pi^*(x_w|y)}{\pi_{\text{ref}}(x_w|y)} + \beta \log Z(y)\right)}{\exp\left(\beta \log \frac{\pi^*(x_w|y)}{\pi_{\text{ref}}(x_w|y)} + \beta \log Z(y)\right) + \exp\left(\beta \log \frac{\pi^*(x_l|y)}{\pi_{\text{ref}}(x_l|y)} + \beta \log Z(y)\right)} \nonumber\\
    &=\frac{1}{1+\exp\left(\beta \log \frac{\pi^*(x_l|y)}{\pi_{\text{ref}}(x_l|y)}-\beta \log \frac{\pi^*(x_w|y)}{\pi_{\text{ref}}(x_w|y)} + (\beta \log Z(y) - \beta \log Z(y))\right)} \nonumber\\
    &=\frac{1}{1+\exp\left(\beta \log \frac{\pi^*(x_l|y)}{\pi_{\text{ref}}(x_l|y)}-\beta \log \frac{\pi^*(x_w|y)}{\pi_{\text{ref}}(x_w|y)}\right)} \nonumber\\
    &= \sigma\left(\beta \log \frac{\pi^*(x_w|y)}{\pi_{\text{ref}}(x_w|y)} - \beta \log \frac{\pi^*(x_l|y)}{\pi_{\text{ref}}(x_l|y)}\right).
    \label{response-conditioned-dpo-derive}
\end{align}

Since we cannot directly access $\pi(x|y)$, we reparameterize it using Bayes' theorem for conditional probability decomposition, transforming Eqn.~\ref{response-conditioned-dpo-derive} into a joint probability formulation as follows:

\begin{align}
    p^*(x_w\succ x_l|y)&= \sigma\left(\beta \log \frac{\pi^*(x_w|y)}{\pi_{\text{ref}}(x_w|y)} - \beta \log \frac{\pi^*(x_l|y)}{\pi_{\text{ref}}(x_l|y)}\right)\nonumber\\
    &= \sigma\left(\beta \log \frac{\frac{\pi^*(x_w, y)}{\pi^*(y)}}{\frac{\pi_{\text{ref}}(x_w, y)}{\pi_{\text{ref}}(y)}} - \beta \log \frac{\frac{\pi^*(x_l, y)}{\pi^*(y)}}{\frac{\pi_{\text{ref}}(x_l, y)}{\pi_{\text{ref}}(y)}}\right)\nonumber\\
    &= \sigma\left(\beta \log \frac{\pi^*(x_w, y)}{\pi_{\text{ref}}(x_w, y)} - \beta \log \frac{\pi^*(x_l, y)}{\pi_{\text{ref}}(x_l, y)}\right).
    \label{eqn:response-conditioned-dpo-derive-joint-prob}
\end{align}

Finally, we derive the loss function for \ourdpo{} for a parameterized policy $\pi_\theta$ using the response-conditioned preference dataset $\mathcal{D}_{Rc} = \{(x_w, x_l, y)\}$ by applying maximum likelihood estimation:

\begin{align}
    \mathcal{L}_{DPO}^{Rc}(\pi_\theta; \pi_\text{ref}) = -\mathbb{E}_{(x_w, x_l, y) \sim \mathcal{D}_{Rc}} \big[\log\sigma\left(\beta \log \frac{\pi_\theta(x_w, y)}{\pi_{\text{ref}}(x_w, y)} - \beta \log \frac{\pi_\theta(x_l, y)}{\pi_{\text{ref}}(x_l, y)}\right)\big].
    \label{eqn:rc-dpo-loss-origin}
\end{align}
\clearpage\twocolumn

\section{The Implementation Details of Preliminary Explorations}
\label{apx:preliminary-implement}

\subsection{Training Dataset and Models}
\label{apx:preliminary_training_data_and_models}

{\bf Training Dataset.} Similar to the data processing approach used by~\citet{yuan2024following}, we extract the first turn of English conversations from the OpenAssistant dataset~\citep{kopf2024openassistant} and use it as our complete dataset. Based on their human annotated ranking, we label rank $0$ as "chosen" ($y_w$) and rank $1$ as "rejected" ($y_l$), resulting in the complete dataset $\mathcal{D}$, that each example is a triple $(x^{(i)}, y_w^{(i)}, y_l^{(i)})$.

For training of different models and subsequent experimental analysis, we first divide $\mathcal{D}$ into a training dataset $\mathcal{D}_{train}$, which contains $90\%$ of the data for model training, and an evaluation dataset $\mathcal{D}_{eval}$, comprising the remaining $10\%$ for model evaluation. Next, we further divide $\mathcal{D}_{train}$ into two subsets: a supervised fine-tuning (SFT) training dataset $\mathcal{D}_{sft}$, containing $30\%$ of $\mathcal{D}_{train}$, and a reward model (RM) training dataset $\mathcal{D}_{rm}$, containing the remaining $70\%$ of $\mathcal{D}_{train}$. For all reward models, we incorporate $\mathcal{D}_{rm}$ into the training process to ensure the fundamental semantic comprehension ability of the trained reward models. For example, in Section \ref{subsec:length-following-ease-learn}, each reward model is trained using the variant dataset ($\text{LIFT-plus}_2^{reverse}$ / $\text{LIFT-plus}_2^{noreverse}$ / $\text{LIFT-plus}_2^{empty}$) combined with $\mathcal{D}_{rm}$.

For the {\em Base} models, we first apply SFT using $\mathcal{D}_{sft}$, followed by reward model training on $\mathcal{D}_{rm}$, For the {\em Instruct} models, we directly train the reward model using $\mathcal{D}_{rm}$ without prior supervised fine-tuning.

{\bf Training Models.} In the preliminary explorations, we employ three groups of models --- Qwen2-1.5B-Base with Qwen2-1.5B-Instruct, Qwen2.5-7B-Base with Qwen2.5-7B-Instruct, and Llama-3.1-8B-Base with Llama-3.1-8B-Instruct --- to verify the broad applicability of our analytical results. However, in Section \ref{subsec:length-following-ease-learn}, we exclude the Qwen2.5-7B-Base and Qwen2.5-7B-Instruct models from our analysis for simplicity. 

\subsection{The Construction Process of $\mathcal{D}_{eval}^q$}
\label{apx:construct-D_eval^q}
To fairly and accurately evaluate the semantic comprehension capabilities of the reward model, we establish a refined quality evaluation dataset $\mathcal{D}_{eval}^q$ through an automated transformation of $\mathcal{D}_{eval}$, leveraging advanced language models \citep{dubois2024alpacafarm, zheng2023judging, chiang2023vicuna}. To ensure the reliability of the generated dataset, we did not directly adopt GPT-4o’s outputs, as GPT-4o also exhibits length bias. Instead, we employed a rigorous algorithm that augments simple model generation with programmatic verification, thereby transforming the process into generation $+$ verification. 

\begin{figure*}[!ht]
    \centering
    \includegraphics[width=\textwidth]{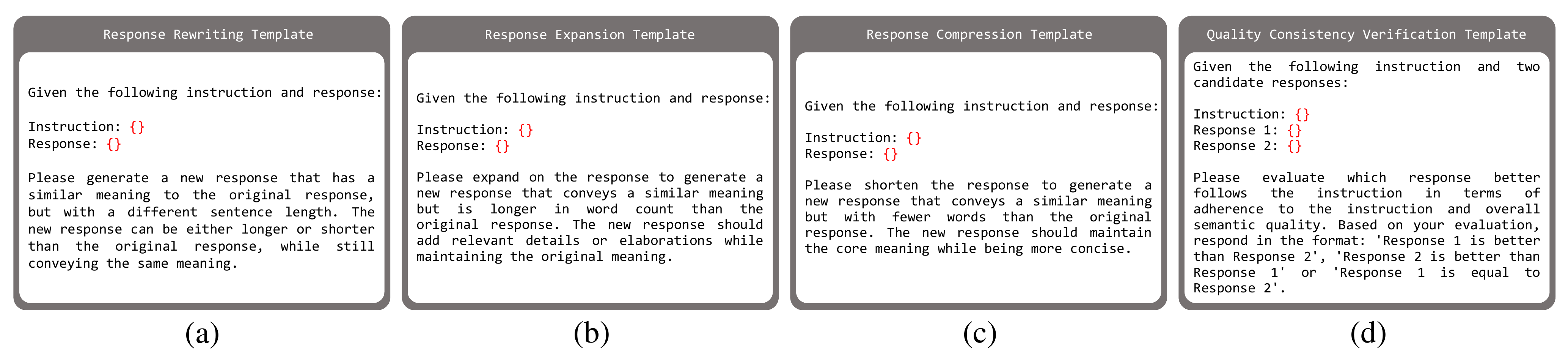}
    \caption{{\bf (a).} Response rewriting template for the chosen response $y_w^{(i)}$; {\bf (b).} Response expansion template for the rejected response $y_l^{(i)}$; {\bf (c).} Response compression template for the rejected response $y_l^{(i)}$; {\bf (d).} Quality consistency verification template for assessing the quality consistency between the rewritten and the original responses.}
    \label{fig:gpt-rewrite-prompt-template}
\end{figure*}

Specifically, First, for each triplet $(x^{(i)},\allowbreak y_w^{(i)},\allowbreak y_l^{(i)})$ in $\mathcal{D}_{eval}$, we employ GPT-4o~\citep{hurst2024gpt} with the {\em Response Rewriting Template} (Figure~\ref{fig:gpt-rewrite-prompt-template}(a)) to generate two alternative responses, $y_w^{(i), 1}$ and $y_w^{(i), 2}$,  based on the original prompt $x^{(i)}$ and response $y_w^{(i)}$. The semantic equivalence between the generated response $y_w^{(i), 1}$ (or $y_w^{(i), 2}$) and the original response $y_w^{(i)}$ is verified using the {\em  Quality Consistency Verification Template} in Figure~\ref{fig:gpt-rewrite-prompt-template}(d).

Subsequently, we process $y_l^{(i)}$ through GPT-4o to generate two alternative responses, $y_l^{(i), 1}$ and $y_l^{(i), 2}$ , based on the original prompt $x^{(i)}$. These generated responses must satisfy two criteria: (1) maintain semantic equivalence with $y_l^{(i)}$, verified using the {\em Quality Consistency Verification Template}, and (2) exhibit specific length relationships with their counterparts, namely $|y_l^{(i), 1}| > |y_w^{(i), 1}|$ and $|y_l^{(i), 2}| < |y_w^{(i), 2}|$, verified using programmatic evaluation. Given GPT-4o's limitations in direct length control, we implement a separate generation process utilizing the {\em Response Expansion Template} and {\em Response Compression Template} (Figure~\ref{fig:gpt-rewrite-prompt-template}(b) and~\ref{fig:gpt-rewrite-prompt-template}(c)) to obtain $y_l^{(i), 1}$ and $y_l^{(i), 2}$, respectively. 

The resulting quality evaluation dataset is formulated as $\mathcal{D}_{eval}^q = \{(x^{(i)}, y_w^{(i), 1}, y_l^{(i), 1})\} \cup \{(x^{(i)}, y_w^{(i), 2}, y_l^{(i), 2})\}$. The complete prompting templates for response generation and quality consistency verification are presented in Figure \ref{fig:gpt-rewrite-prompt-template}.

\subsection{The Construction Process of $\mathcal{D}_{eval}^l$}
\label{apx:construct-D_eval^l}

To accurately and efficiently evaluate the reward model's adherence to length instructions, we establish the length evaluation dataset $\mathcal{D}_{eval}^l$. Each triplet $(x_l^{(i)}, y_w^{(i), l}, y_l^{(i), l})$ in $\mathcal{D}_{eval}^l$ consists of a length instruction $x_l^{(i)} = \text{\em length constraint} + x^{(i)}$ and two semantically equivalent responses that differ only in length: $y_w^{(i), l}$ adheres to the length constraint specified in $x_l^{(i)}$, while $y_l^{(i), l}$ violates it\footnote{The detailed definition of the {\em length constraint} can be found in Appendix \ref{apx:LIFT-decomposition}.}. To optimize computational efficiency, we derive $\mathcal{D}_{eval}^l$ from the previously constructed $\mathcal{D}_{eval}^q$. Specifically, for the responses in $\mathcal{D}_{eval}^q$, we already ensure that $y_w^{(i), 1}$ and $y_w^{(i), 2}$ maintain semantic similarity during the construction of $\mathcal{D}_{eval}^q$. Additionally, the stochastic nature of GPT-4o's response generation inherently produces responses of varying lengths, with $|y_w^{(i), 1}|$ and $|y_w^{(i), 2}|$ typically differing. Leveraging this property, we select these responses ($y_w^{(i), 1}$ and $y_w^{(i), 2}$) as candidates for $x_l^{(i)}$. Specifically, we set the {\em word\_num} parameter in $x_l^{(i)}$ within the range $[|y_w^{(i), 1}|, |y_w^{(i), 2}|]$ (assuming $|y_w^{(i), 1}| < |y_w^{(i), 2}|$, $x_l^{(i)} = \text{\em less length constraint} + x^{(i)}$). Consequently, we assign   $y_w^{(i), l} = y_w^{(i), 1}$ as the chosen response and $y_l^{(i), l} = y_w^{(i), 2}$ as the rejected response. Finally, the constructed length evaluation dataset is $\mathcal{D}_{eval}^l = \{(x_l^{(i)}, y_w^{(i), l}, y_l^{(i), l})\}$.

\subsection{The Extension Process of LIFT}
\label{apx:LIFT-decomposition}

In LIFT \citep{yuan2024following}, the authors propose a straightforward method for constructing a length instruction dataset. In Section \ref{subsec:length-following-ease-learn}, we will examine the limitations of this method. The LIFT approach can be divided into two types: the first part $\text{LIFT}_1$ constructs length instruction $x_l^{(i)}$ that both responses ($y_w^{(i)}$ and $y_l^{(i)}$) satisfy the length constraint specified in $x_l^{(i)}$; the second part $\text{LIFT}_2$ constructs length instruction $x_l^{(i)}$ that only one response ($y_w^{(i)}$ or $y_l^{(i)}$) satisfies. In cases where the rejected response $y_l^{(i)}$ adheres to the length instruction but the chosen response $y_w^{(i)}$ does not, this can lead to a reversal of the original preference order, which fundamentally undermines the semantic quality of the data. Furthermore, LIFT only considers a maximum length constraint, i.e., {\em "less length constraint"}, without accounting for a minimum length constraint, i.e., {\em "more length constraint"}. Therefore, we made a slight improvement to LIFT by adding an {\em "more length constraint"}, resulting in LIFT-plus:

\begin{equation}
\small
    \begin{aligned}
        \text{LIFT-plus}_{less} & = \text{LIFT} = \text{LIFT}_1 \cup \text{LIFT}_2 \\
        & = \{(x_{l, less}^{(i)}, y_w^{(i)}, y_l^{(i)})\} \cup \{(x_{l, less}^{(i)}, y_l^{(i)}, y_w^{(i)})\}, \\
        x_{l, less}^{(i)} & = \text{\em less length constraint} + x^{(i)}, \\
        \text{LIFT-plus}_{more} & = \{(x_{l, more}^{(i)}, y_w^{(i)},  y_l^{(i)})\} \cup \{(x_{l, more}^{(i)}, y_l^{(i)},  y_w^{(i)})\}, \\
        x_{l, more}^{(i)} & = \text{\em more length constraint} + x^{(i)}, \\
        \text{LIFT-plus} & = \text{LIFT-plus}_{less} \cup \text{LIFT-plus}_{more}. 
    \end{aligned}
\end{equation}

\noindent
where less length constraint is "Answer the following instruction using $\{word\_num\}$ words or less.", and more length constraint is "Answer the following instruction using $\{word\_num\}$ words or more."
In this article, we refer to $x_{l, less}^{(i)}$ as the "or less" length instruction and $x_{l, more}^{(i)}$ as the "or more" length instruction. Similarly, LIFT-plus can also be decomposed into two parts: $\text{LIFT-plus}_1$ and $\text{LIFT-plus}_2$ as follows:

\begin{equation}
\small
    \begin{aligned}
        \text{LIFT-plus} &= \text{LIFT-plus}_{1} \cup \text{LIFT-plus}_{2} \\
        &= \{(x_l^{(i)}, y_w^{(i)}, y_l^{(i)})\} \cup \{(x_l^{(i)}, y_l^{(i)}, y_w^{(i)})\}, \\
        x_l^{(i)} &= \text{\em length constraint} + x^{(i)}.
    \end{aligned}
\end{equation}

\noindent
where length constraint is less length constraint or more length constraint.
In the first part $\text{LIFT-plus}_1$, the length instruction $x_l^{(i)}$ is satisfied to both responses. As a result, it may degrade into a length-agnostic instruction, where the model disregards the added length constraint and focuses solely on the original prompt $x^{(i)}$.
Therefore, the key component that effectively facilitates adherence to length instructions is the second part, $\text{LIFT-plus}_2$, where only one response ($y_w^{(i)}$ or $y_l^{(i)}$) satisfies $x_l^{(i)}$. 

Furthermore, we divide $\text{LIFT-plus}_2$ into two subsets: $\text{LIFT-plus}_2^{reverse} = \{(x_l^{(i)}, y_l^{(i)}, y_w^{(i)})\}$, which reverses the original preference order between chosen and rejected responses due to the length constraint in $x_l^{(i)}$, and $\text{LIFT-plus}_2^{noreverse} = \{(x_l^{(i)}, y_w^{(i)}, y_l^{(i)})\}$, which preserves the original preference order. In addition, based on $\text{LIFT-plus}_2^{noreverse}$, we construct a dataset purely centered on length instructions, denoted as $\text{LIFT-plus}_2^{empty} = {(x_e^{(i)}, y_w^{(i)}, y_l^{(i)})}$, where $x_e^{(i)} =$ {\em length constraint} $+$ {\em ["empty prompt"]}. This dataset isolates length instructions from other semantic content and only focuses on the length instructions themselves. Here, $\text{LIFT-plus}_2^{empty}$ is specifically designed to investigate the impact of semantic prompts on length instruction following in Section \ref{subsec:length-following-ease-learn}.

\section{The Detailed Results and Analysis of Preliminary Explorations}
\label{appendix:additional-result-of-preliminary-experiments}

\subsection{Length Bias Indeed Exists}
\label{appendix:lenght-bias-in-model}
Although the responses $y_w^{(i)}$ and $y_l^{(i)}$ in both $\mathcal{D}_{eval}^e$ and $\mathcal{D}_{eval}^r$ are semantically misaligned with their respective prompt, we observe that the reward models still achieve relatively high accuracies on these evaluation datasets. Specifically, the results for the three groups of models on $\mathcal{D}_{eval}^e$ and $\mathcal{D}_{eval}^r$, as shown in Table~\ref{tab:baseline-rm-original-empty-prompt-result}, indicate that nearly all reward models achieve an accuracy of $60\%$ or higher, which is very close to the accuracy observed on the original evaluation dataset $\mathcal{D}_{eval}$. This suggests that, even in the absence of the prompt, the reward models exhibit a significant bias toward the chosen responses.

To further investigate whether this bias primarily arises from response length, we first analyze the consistency of model evaluation results between $\mathcal{D}_{eval}^e$ (or $\mathcal{D}_{eval}^r$) and $\mathcal{D}_{eval}$, focusing on how varying prompts affect the selection of the same response. The results, presented in Table~\ref{tab:baseline-rm-original-empty-prompt-result}, show that in over $85\%$ of cases, the models select the same response regardless of the prompt. This reinforces the observation that the reward models' preference is not primarily driven by the prompts themselves but rather by other response-related factors.

Next, we plot the relationship between the response length and the corresponding reward score of the reward models for $\mathcal{D}_{eval}$, $\mathcal{D}_{eval}^e$, and $\mathcal{D}_{eval}^r$, as shown in Figure~\ref{fig:length-score-ranges-baseline}. To facilitate comparison, we normalize the reward scores of both models (Qwen2-1.5B-Instruct and Llama-3.1-8B-Instruct) to the same range. The results reveal a strong linear correlation between the response length and the reward score across all three evaluation datasets. Specifically, as response length increases, the model's reward score also rises, indicating that response length plays a significant role in the model's assessment of response quality.

We further conducted extra experiments that analyzing the relationship between the accuracy of reward models on $\mathcal{D}_{eval}$ and the length difference between the chosen and rejected responses. The results are demonstrated in Table~\ref{tab:original-rm-result-diff-length}. It can be observed that as the length difference increases, the accuracy also increases. This clearly demonstrates the presence of length bias in the Baseline models.

\begin{table*}[!ht]
    \centering
    \caption{Accuracy of different reward models on different subsets of $\mathcal{D}_{eval}$ partitioned by length difference between chosen and rejected responses}
    \begin{tabular}{lcccccc}
    \toprule
        Model & $\leq$ -100 & -100 $\sim$ -50 & -50 $\sim$ 0 & 0 $\sim$ 50 & 50 $\sim$ 100 & $\geq$ 100 \\ 
        \midrule
        Qwen2-1.5B-Instruct & 9.10 & 6.06 & 38.96 & 84.40 & 89.79 & 95.52 \\ 
        \midrule
        Qwen2.5-7B-Instruct & 18.19 & 15.16 & 29.87 & 81.65 & 91.84 & 88.46 \\ 
        \midrule
        Llama-3.1-8B-Instruct & 18.19 & 9.10 & 40.26 & 67.89 & 79.59 & 88.06 \\
        \bottomrule
    \end{tabular}
    \label{tab:original-rm-result-diff-length}
\end{table*}

\subsection{Length Bias in Evaluation Dataset}
\label{appendix:length-bias-in-data}

We conducted further analysis on the generated evaluation dataset $\mathcal{D}_{eval}^q$. Specifically, we recruited $20$ volunteers, each assigned $10$ randomly sampled data from $\mathcal{D}_{eval}^q$. The volunteers were instructed to focus solely on the semantic quality of the responses, disregarding factors such as length or format. We then collected their preferences regarding the data. Upon aggregating the results, we found that human preferences are $90.5\%$ consistent with $\mathcal{D}_{eval}^q$. This strongly supports the validity of our generated datasets.

\subsection{Length Instructions Are Easily Learned}
\label{appendix:length-following-ease-learn}

The results for the Qwen2-1.5B and Llama-3.1-8B models trained on the three variant datasets ($\text{LIFT-plus}_2^{reverse}$, $\text{LIFT-plus}_2^{noreverse}$, and $\text{LIFT-plus}_2^{empty}$) are presented in Table~\ref{rm-length-prompt-result-reverse/unreverse/empty_prompt}. As observed, most reward models achieve similar or even lower accuracy on $\mathcal{D}_{eval}^q$ compared to Baseline models in Table~\ref{tab:rm-result}, while their accuracy on $\mathcal{D}_{eval}^l$ remains relatively high. These results suggest that, although reward models are trained on diverse length instruction datasets, they all primarily learn to capture response length information rather than balancing adherence to length instructions with attention to the semantic content of the prompt.

\begin{figure*}[!ht]
    \centering
    \subfigure[Qwen2-1.5B-Instruct]{\includegraphics[width=0.45\textwidth]{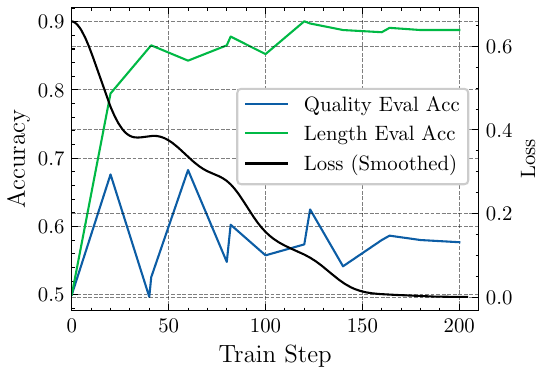}}
    \subfigure[Llama-3.1-8B-Instruct]{\includegraphics[width=0.452\textwidth]{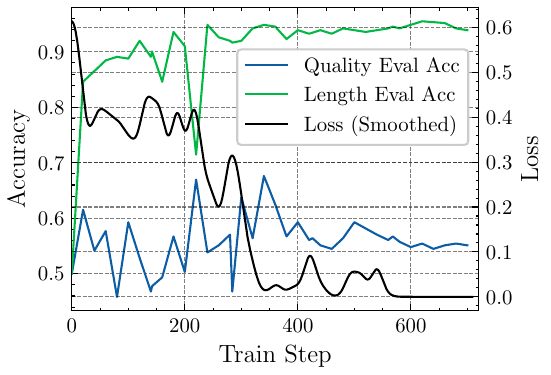}}
    \caption{The trajectories of Quality Eval Acc, Length Eval Acc, and Training Loss (Smoothed) for different reward models trained on $\text{LIFT-plus}_2^{reverse}$ across training steps.}
    \label{fig:length-quality-acc-train-loss-train-step}
\end{figure*}

\begin{table*}[!ht]
    \centering
    \caption{Evaluation results of different reward models (LIFT-plus variants) on quality ($\mathcal{D}_{eval}^q$) and length  ($\mathcal{D}_{eval}^l$) evaluation datasets.}
    \begin{tabular}{llcc}
    \toprule
        Model & Variant & Quality Eval Acc ($\%$) & Length Eval Acc ($\%$) \\ 
        \midrule
        \multirow{3}{*}{Qwen2-1.5B-Base} & $\text{LIFT-plus}_2^{reverse}$ & 58.78 & 85.58 \\ 
        & $\text{LIFT-plus}_2^{noreverse}$ & 59.41 & 87.78 \\
        & $\text{LIFT-plus}_2^{empty}$ & 58.33 & 89.71 \\
        \midrule
        \multirow{3}{*}{Qwen2-1.5B-Instruct} & $\text{LIFT-plus}_2^{reverse}$ & 60.11 & 86.54 \\ 
        & $\text{LIFT-plus}_2^{noreverse}$ & 63.17 & 85.21 \\
        & $\text{LIFT-plus}_2^{empty}$ & 61.02 & 85.53 \\
        \midrule
        \multirow{3}{*}{Llama-3.1-8B-Base} & $\text{LIFT-plus}_2^{reverse}$ & 51.34 & 91.32 \\ 
        & $\text{LIFT-plus}_2^{noreverse}$ & 57.71 & 88.14 \\
        & $\text{LIFT-plus}_2^{empty}$ & 50.27 & 88.78 \\
        \midrule
        \multirow{3}{*}{Llama-3.1-8B-Instruct} & $\text{LIFT-plus}_2^{reverse}$ & 60.11 & 95.19 \\ 
        & $\text{LIFT-plus}_2^{noreverse}$ & 56.65 & 92.31 \\
        & $\text{LIFT-plus}_2^{empty}$ & 57.97 & 90.71 \\
        \bottomrule
    \end{tabular}
    \label{rm-length-prompt-result-reverse/unreverse/empty_prompt}
\end{table*}
To further illustrate how reward models quickly overfit to length instructions during training, thereby impeding semantic learning, we visualize the accuracy trajectories of Qwen-2-1.5B-Instruct and Llama-3.1-8B-Instruct on  $\mathcal{D}_{eval}^q$ and $\mathcal{D}_{eval}^l$ as training steps increase (see Figure \ref{fig:length-quality-acc-train-loss-train-step}). The results show that as training progresses, the model accuracy on $\mathcal{D}_{eval}^l$ increases rapidly, whereas the accuracy on $\mathcal{D}_{eval}^q$ improves at a much slower rate. Moreover, once the accuracy on $\mathcal{D}_{eval}^l$ reaches its peak, the accuracy on $\mathcal{D}_{eval}^q$ also begins to decline and gradually plateaus. This tendency strongly supports the conclusion that the reward models quickly prioritize learning length instructions, achieving high accuracy on length-related tasks, but at the expense of semantic quality. The slow improvement in the accuracy on $\mathcal{D}_{eval}^q$ further suggests that the reward models struggle to learn the necessary semantic understanding, as they are overly focused on conforming to length constraints.

\section{Additional Experimental Results and Analysis}
\label{apx:additional-experimental-results}

\subsection{The Evaluation Details of DPO Models}
\label{apx:dpo-eval-details}
Similar to Appendix \ref{apx:LIFT-decomposition}, due to the length instruction $x_l$ in the AlpacaEval-LI benchmark provided by LIFT~\citep{yuan2024following} includes only the "or less" length instruction, it is necessary to incorporate the "or more" length instruction. Therefore, we extend the AlpacaEval-LI benchmark to include both the "or less" and "or more'' length instructions. Specifically, in AlpacaEval-LI, $word\_num$ is set based on the shortest response length for the original prompt $x$ across three advanced models --- GPT-4~\citep{achiam2023gpt}, Claude3-Opus, and Mistral Large --- which we assume approximates the median response length for each prompt. This suggests that well-reasoned, detailed responses may exceed $word\_num$, while concise yet valid responses may fall below it. Therefore, we improve the AlpacaEval-LI benchmark by modifying the "or less" in $x_l$ to "or more", specifically replacing "Answer the following instruction using \{{\em word\_num}\} or less" with "Answer the following instruction using \{{\em word\_num}\} or more", while keeping all other aspects unchanged. This results in the expanded AlpacaEval-LI-plus-more benchmark, and the original AlpacaEval-LI benchmark is referred to as AlpacaEval-LI-plus-less benchmark. In cases where there is no ambiguity, we refer to AlpacaEval-LI-plus-less and AlpacaEval-LI-plus-more benchmark as AlpacaEval-LI-plus benchmark.

\begin{figure*}[!ht]
    \centering
    \includegraphics[width=\textwidth]{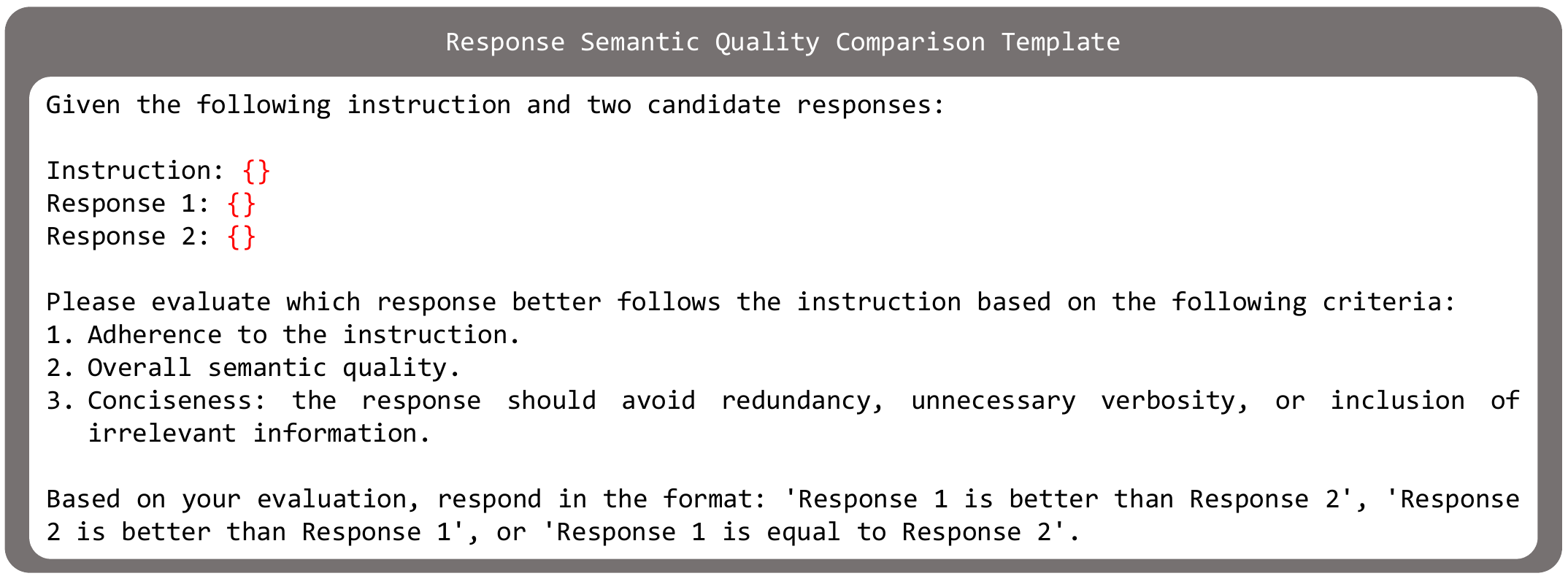}
    \caption{Response comparison template for evaluating the semantic quality between responses.}
    \label{fig:gpt-eval-response-prompt-template}
\end{figure*}

For the DPO evaluation metrics, as described in Section \ref{subsec:experiment-settings}, we use automated assessment by leveraging GPT-4o~\citep{hurst2024gpt} to label each pair of responses as {\em win}, {\em tie}, or {\em lose} based on semantic quality. The prompt used to guide GPT-4o's evaluation is similar to the one provided in Appendix~\ref{apx:construct-D_eval^q}, as shown in Figure \ref{fig:gpt-eval-response-prompt-template}. Specifically, we evaluate each DPO model's performance by comparing it with its corresponding SFT/{\em Instruct} model using three metrics: {\em Length Acc} refers to the accuracy of the model-generated response length that satisfies the length constraint specified in the given length instruction $x_l$ in AlpacaEval-LI-plus. {\em Length Win Ratio} denotes the proportion of cases in which the model-generated response is labeled as {\em win} compared to the response generated by the SFT/{\em Instruct} models, under the given length instruction $x_l$. {\em Quality Win Ratio} measures the proportion of cases in which the model-generated response is labeled as {\em win} compared to the response generated by the SFT/{\em Instruct} models under the given original prompt $x$. {\em Length Win Ratio} and {\em Quality Win Ratio} evaluate the semantic quality of responses generated by the DPO models under different prompts.

\subsection{The Results of Reward Models}
\label{additional-results-rm}

\begin{table}
    \centering
    \small
    \caption{\LEA{} of different reward models on length ($\mathcal{D}_{eval}^l$) evaluation datasets.}
    \label{tab:rm-result-length}
    \setlength{\tabcolsep}{5pt}
    \begin{tabular}{lccc}
        \toprule
        Model & Baseline & LIFT-plus & Rc-RM \\
        \midrule
        Qwen2-1.5B-Base & 52.41 & \textbf{87.18} & 86.22 \\
        Qwen2-1.5B-Instruct & 51.77 & \textbf{86.54} & 84.61 \\
        \midrule
        Qwen2.5-7B-Base & 52.24 & 84.94 & \textbf{88.14} \\
        Qwen2.5-7B-Instruct & 52.88 & 83.97 & \textbf{92.31} \\
        \midrule
        Llama-3.1-8B-Base & 51.92 & \textbf{91.32} & 88.46 \\
        Llama-3.1-8B-Instruct & 49.04 & \textbf{95.19} & 93.27 \\
        \bottomrule
    \end{tabular}
\end{table}

{\bf Rc-RM effectively follows length instructions.} In addition to mitigating length bias, \ourrm{} significantly enhances the ability to follow length instructions, as demonstrated by the {\em Length Eval Acc} results in Table \ref{tab:rm-result-length}. Specifically, \ourrm{} achieves accuracy comparable to LIFT-plus and surpasses it in nearly half of the models, with gains of $3.2\%$ on Qwen2.5-7B-Base, and $8.34\%$ on Qwen2.5-7B-Instruct. 

Taking "or less" length instruction as a representative experiment, we evaluate the score consistency of reward models with respect to length instructions by constructing a sequence of length instructions $x_l^{(i, j)}, j = \{1, 2, ..., 8\}$, with increasing {\em word\_num}. Specifically, we first select all instances $(x^{(i)}, y_w^{(i)}, y_l^{(i)})$ in $\mathcal{D}_{eval}$ where $|y_w^{(i)}| < |y_l^{(i)}|$, referring to this subset as $\mathcal{D}_{eval}^{less}$. For each $x^{(i)}$ in $\mathcal{D}_{eval}^{less}$, we then construct the following sequence of {\em word\_num} values ($L_{wn}$) to generate the corresponding sequence of length instructions $x_l^{(i, j)}$, and refer to the entire constructed evaluation dataset as $\mathcal{D}_{eval}^{mls}$:

\begin{equation}
    \begin{aligned}
        L_{wn} =&[\,l_w-2T,\, l_w-T,\, l_w,\, l_w+L,\\
                    &l_w+2L,\, l_l,\, l_l+T,\, l_l+2T\,], \\
        \mathcal{D}_{eval}^{mls} =&\bigl\{(x_{l}^{(i,j)},\, y_{w}^{(i)},\, y_{l}^{(i)})\bigr\},\quad j\in\{1,2,\dots,8\},
    \end{aligned}
\end{equation}

\noindent
where $x_{l}^{(i,j)} =$ "Answer the following instruction using $\{word\_num=L_{wn}^{(j)}\}$ words or less." $+\ x^{(i)}$,  $T = 10$, $L = (l_l - l_w) / 3$, $l_w = |y_w|$, and $l_l = |y_l|$. For the sequence of $(x_l^{(i, j)}, y_w^{(i)}, y_l^{(i)}), j=\{1, 2, ..., 8\}$ in $\mathcal{D}_{eval}^{mls}$, the ideal performance of the reward model should be as follows: Initially, when $word\_num < l_w$, the length constraint of $x_l^{(i, j)}$ is invalid for both $y_w^{(i)}$ and $y_l^{(i)}$, the difference in reward scores between $y_w^{(i)}$ and $y_l^{(i)}$ primarily reflect the semantic difference in the original data $(x^{(i)}, y_w^{(i)}, y_l^{(i)})$. When $word\_num = l_w$, the length constraint of $x_l^{(i, j)}$ aligns with $y_w^{(i)}$ but not $y_l^{(i)}$, resulting in a greater difference between $y_w^{(i)}$ and $y_l^{(i)}$ compared to their original semantic difference. When $word\_num = l_l$, the length constraint of $x_l^{(i, j)}$ becomes valid for both $y_w^{(i)}$ and $y_l^{(i)}$, and the difference reverts to those derived from the original semantics. Therefore, the difference in the predicted scores of the reward model for $y_w^{(i)}$ and $y_l^{(i)}$ should initially increase with the rise in $word\_num$, then decrease, eventually returning to the original semantic difference. 

\begin{figure}
    \centering
    \vspace{-0.2cm}
    \subfigure[Qwen2-1.5B-Instruct]{\includegraphics[width=0.23\textwidth]{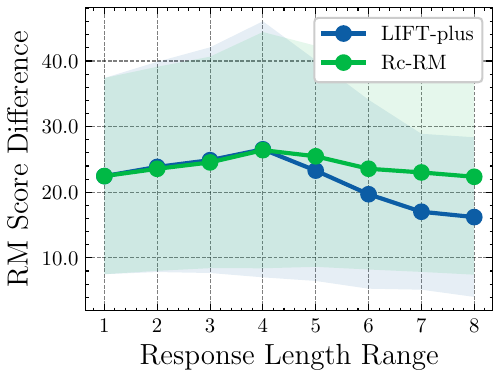}}
    \subfigure[Llama-3.1-8B-Instruct]{\includegraphics[width=0.23\textwidth]{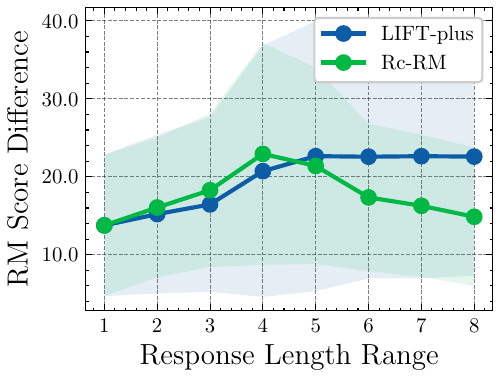}}
    \caption{Comparison of score differences between LIFT-plus and Rc-RM on $\mathcal{D}_{eval}^{mls}$ under varying {\em word\_num} constraints. An ideal reward model should show an initial increase in score difference followed by a return to its initial value.}
    \vspace{-0.2cm}
    \label{fig:multi-length-bias-multi-wordnum-rm-score-diff-2}
\end{figure}

The results, as shown in Figure \ref{fig:multi-length-bias-multi-wordnum-rm-score-diff-2}(c) and \ref{fig:multi-length-bias-multi-wordnum-rm-score-diff-2}(d), demonstrate that both LIFT-plus and \ourrm{} exhibit an initial increase followed by a decrease in reward score differences as {\em word\_num} increases. However, LIFT-plus fails to return to the original difference after {\em word\_num} increases, instead remaining at a level higher or lower than that observed when {\em word\_num} $< |y_w|$. This indicates that LIFT-plus overfits to the length instructions, thereby neglecting the original semantic instructions. In contrast, \ourrm{} accurately reflects the expected behavior: the score difference increases with the rise in {\em word\_num}, then decreases, eventually returning to the original semantic difference. Moreover, \ourrm{} exhibits minimal score fluctuation at both extremes -- when length constraints are invalid for both responses and when they are valid for both responses. This robust performance provides strong evidence for the effectiveness of our approach in simultaneously adhering to both length and original semantic instructions.

\subsection{The Results of Qwen2.5-1.5B Reward Models}
\label{qwen2.5-1.5b-results}

This subsection mainly presents the RM experimental results for Qwen2.5-1.5B and some unreasonable experimental phenomena. As show in Table \ref{tab:rm-result-qwen2.5-1.5b}, for Qwen2.5-1.5B-Base, the results are similar to those to Section \ref{subsec:rm-results}. \ourrm{} not only achieves higher performance in {\em Quality Eval Acc}, being $7.69\%$ higher than LIFT-plus and $6.41\%$ higher than ODIN, but also performs almost equally with LIFT-plus in {\em Length Eval Acc}. For Qwen2.5-1.5B-Instruct, \ourrm{} still outperforms LIFT-plus by $3.21\%$ and ODIN by $1.61\%$ in {\em Quality Eval Acc}. However, the results for LIFT-plus are somewhat unreasonable: its {\em Length Eval Acc} is only $67.63\%$, $12.5\%$ lower than \ourbtmodel{}, while its {\em Quality Eval Acc} is $10.9\%$ higher than Baseline.

\begin{table*}[!ht]
    \centering
    \small
    \caption{Evaluation results of reward models on quality ($\mathcal{D}_{eval}^q$) and length ($\mathcal{D}_{eval}^l$) evaluation datasets (Qwen2.5-1.5B).}
    \begin{tabular}{ccccccccc}
        \toprule
        \multirow{2}{*}{Metrics} & \multicolumn{4}{c}{Qwen2.5-1.5B-Base} & \multicolumn{4}{c}{Qwen2.5-1.5B-Instruct} \\
        \cmidrule(lr){2-5} \cmidrule(lr){6-9} & Baseline & LIFT-plus & ODIN & Rc-RM & Baseline & LIFT-plus & ODIN & Rc-RM \\
        \cmidrule(lr){1-1} Quality Eval Acc ($\%$) & 58.97 & 60.26 & 61.54 & \textbf{67.95} & 58.01 & 68.91 & 70.51 & \textbf{72.12} \\
        Length Eval Acc ($\%$) & 51.28 & \textbf{87.82} & 55.45 & 85.89 & 53.53 & 67.63 & 50.00 & \textbf{80.13} \\
        \bottomrule
    \end{tabular}
    \label{tab:rm-result-qwen2.5-1.5b}
\end{table*}

\begin{figure*}[!ht]
    \centering
    \subfigure[Length Eval Acc]{\includegraphics[width=0.48\textwidth]{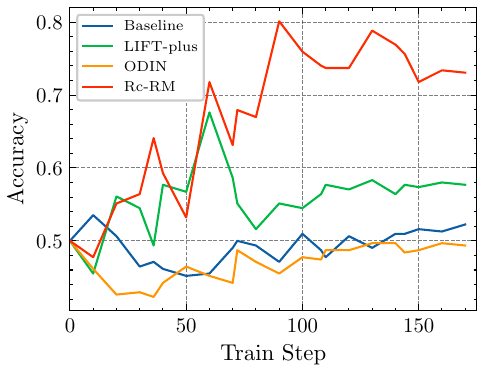}}
    \subfigure[Quality Eval Acc]{\includegraphics[width=0.49\textwidth]{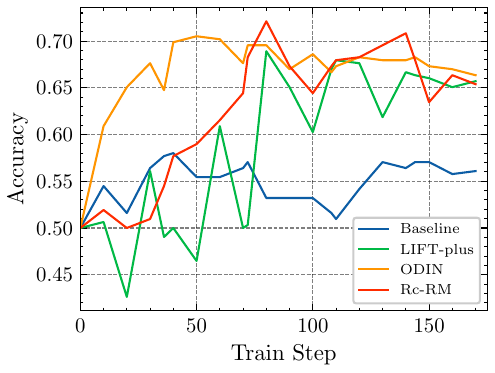}}
    \caption{The trajectories in Length Eval Acc and Quality Eval Acc of Qwen2.5-1.5B-Instruct with training steps.}
    \label{fig:length-quality-acc-train-step-qwen2.5-1.5b-instruct}
\end{figure*}

Furthermore, we plot the {\em Quality Eval Acc} and {\em Length Eval Acc} trajectories for each method across training steps, as shown in Figure \ref{fig:length-quality-acc-train-step-qwen2.5-1.5b-instruct}. It can be observed that in the early stages of training, LIFT-plus's {\em Length Eval Acc} gradually increases, while its {\em Quality Eval Acc} is worse than that of Baseline, which aligns with our analysis in Section \ref{subsec:length-following-ease-learn}. However, after just $60$ steps of training, LIFT-plus's {\em Length Eval Acc} suddenly drops drastically, while {\em Quality Eval Acc} begins to rise significantly. This ultimately results in {\em Length Eval Acc} stabilizing at a relatively low value and {\em Quality Eval Acc} stabilizing at a relatively high value. Since this is a single isolated example, we suspect it may be related to the specific model type and will conduct a more detailed analysis in future work. 

\begin{table*}[!ht]
    \centering
    \small
    \caption{Evaluation results of different DPO models on AlpacaEval-LI-plus-less.}
    \begin{tabular}{ccccccccc}
        \toprule
        \multirow{2}{*}{Metrics} & \multicolumn{4}{c}{Qwen2.5-7B-Base} & \multicolumn{4}{c}{Qwen2.5-7B-Instruct} \\
        \cmidrule(lr){2-5} \cmidrule(lr){6-9} & Baseline & LIFT-plus & R-DPO & Rc-DPO & Baseline & LIFT-plus & R-DPO & Rc-DPO \\
        \cmidrule(lr){1-1} Length Acc (\%) & 5.74 & \textbf{86.78} & 8.35 & 82.04 & 89.65 & \textbf{100} & 90.90 & \textbf{100} \\
        Response Length & 544.47 & 106.87 & 637.44 & 177.40 & 136.75 & 23.56 & 131.17 & 108.24 \\ 
        Length Win Ratio (\%) & 34.04 & 1.37 & 34.54 & \textbf{44.14} & 32.17 & 2.99 & 31.92 & \textbf{50.75} \\ 
        \midrule
        \multirow{2}{*}{Metrics} & \multicolumn{4}{c}{Llama-3.1-8B-Base} & \multicolumn{4}{c}{Llama-3.1-8B-Instruct} \\
        \cmidrule(lr){2-5} \cmidrule(lr){6-9} & Baseline & LIFT-plus & R-DPO & Rc-DPO & Baseline & LIFT-plus & R-DPO & Rc-DPO \\
        \cmidrule(lr){1-1} Length Acc (\%) & 13.59 & \textbf{98.75} & 27.68 & 94.51 & 87.66 & \textbf{100} & 87.03 & \textbf{100} \\
        Response Length & 424.47 & 6.62 & 518.74 & 118.50 & 150.43 & 35.01 & 151.39 & 89.75 \\ 
        Length Win Ratio (\%) & 35.41 & 0.25 & 47.88 & \textbf{64.96} & 43.89 & 1.00 & 43.77 & \textbf{64.71} \\
        \bottomrule
    \end{tabular}
    \vspace{-0.2cm}
    \label{tab:dpo-length-result-or-less}
\end{table*}

\subsection{The Results of DPO Models in AlpacaEval-LI-plus-less}
\label{or-less-results-analysis}

Due to length bias, models tend to generate longer responses, leading to higher {\em Length Acc} on the AlpacaEval-LI-plus-more benchmark, even when the models have no awareness of length instructions. This makes it difficult to fairly compare models' ability to follow length instructions. Therefore, by examining the performance on the AlpacaEval-LI-plus-less benchmark, we can better assess how effectively each method adheres to the specified length limits while still maintaining a high level of semantic quality in the generated responses.

The results on AlpacaEval-LI-plus-less are shown in Tabel~\ref{tab:dpo-length-result-or-less}. As observed, for the {\em Base} models, Baseline models are trained only on $\mathcal{D}_{sft}$ and $\mathcal{D}_{rm}$, and do not demonstrate the ability to follow length instructions. On the other hand, the {\em Instruct} models inherently exhibit a stronger ability to follow length instructions. Consequently, Baseline models trained on the \textit{Instruct} models also acquire a certain degree of length instruction adherence. For R-DPO, since it was trained only on $\mathcal{D}_{rm}$, and according to the {\em Response Length} results on Tables~\ref{tab:dpo-origin-quality-result},~\ref{tab:dpo-length-result-or-less} and~\ref{tab:dpo-length-result-or-more}, it primarily reduces the response length across all instructions (include "or more" length instruction), without explicitly following the length constraints. This indicates that R-DPO does not specifically address length instruction adherence, but simply focuses on minimizing the response length overall. Thus, the main comparison method we focus on here is LIFT-plus.

It is evident that LIFT-plus suffers from a severe "short bias" issue in Appendix~\ref{apx:short-bias-lift}, focusing exclusively on the "or less" length instruction $x_l$ itself without considering the specified {\em word\_num} or the original prompt $x$. This leads to extremely short responses that strictly comply with the length instruction but completely disregard the semantic requirement of the original prompt. Specifically, On Qwen2.5-7B-Base and Qwen2.5-7B-Instruct, the average response lengths generated by LIFT-plus are only $106.87$ and $23.56$, significantly below the average length constraint of $180.23$ in AlpacaEval-LI-plus-less. On Llama-3.1-8B-Instruct, the average response length is $35.01$, and on Llama-3.1-8B-Base, the average response length drops to a mere $6.62$, showing a 111.88-word difference compared to \ourbtmodel{}. This behavior results in LIFT-plus achieving nearly $100\%$ {\em Length Acc}, but with {\em Length Win Ratio} close to $0\%$, which fails to align with the desired semantic quality.

In contrast, \ourdpo{} not only adheres to the "or less" length instruction but also considers the specified {\em word\_num} and original prompt. It strives to maximize the semantic quality of the response within the length constraint. Therefore, compared to the {\em Quality Win Ratio} in Table~\ref{tab:dpo-origin-quality-result}, \ourdpo{} maintains semantic quality even under length constraints. Notably, on Qwen2.5-7B-Base and Llama-3.1-8B-Base, \ourbtmodel{} achieves $82.04\%$ and $94.51\%$ {\em Length Acc}, respectively, while boosting the {\em Length Win Ratio} to $44.14\%$ and $64.96\%$, further enhancing the semantic abilities of SFT models. For both Qwen2.5-7B-Instruct and Llama-3.1-8B-Instruct, \ourbtmodel{} achieves $100\%$ {\em Length Accuracy}, while obtaining {\em Length Win Ratio} of $50.75\%$ and $64.71\%$, respectively. This fully demonstrates the effectiveness of \ourdpo{} in balancing adherence to length instructions with high semantic quality.

\subsection{The Results of DPO Models in AlpacaEval-LI-plus-more}
\label{or-more-results-analysis}
In this subsection, we present the experimental results of the DPO models from Section \ref{subsec:dpo-results} on AlpacaEval-LI-plus-more, as shown in Table \ref{tab:dpo-length-result-or-more}. Similar to Section \ref{subsec:dpo-results}, since R-DPO was not trained on length instruction data, we primarily compare our method with LIFT-plus. First, as seen in Table \ref{tab:dpo-origin-quality-result}, due to the influence of length bias, Baseline model outputs longer responses, which means that it largely satisfies the "or more" length instruction limit without focusing on the length constraint itself. Moreover, a comparison with Table \ref{tab:dpo-length-result-or-more} further confirms it: after adding the "or more" length instruction limit, the length variation in the Baseline responses is not significant (especially for the {\em Instruct} model). At the same time, its {\em Length Win Ratio} demonstrates that compared to the SFT / {\em Instruct} models, the Baseline's responses are more redundant and repetitive, rather than truly improving semantic quality, leading to no significant improvement in {\em Length Win Ratio}. In fact, for all four models, the Baseline's {\em Length Win Ratio} is lower than ours by $17.46\%$, $24.31\%$, $9.61\%$, and $14.71\%$, respectively.

For LIFT-plus, the results are similar to those in Appendix~\ref{apx:short-bias-lift}, where it focuses only on the "or more" length instruction and disregards the semantic requirements of the original prompt and the specific length constraints of the "or more" length instruction. This causes the model to generate excessively long and unnecessary responses, resulting in a high {\em Length Acc} but a decline in {\em Length Win Ratio}. Specifically, for the {\em Instruct} models, due to the strong foundational instruct-following capability, the over-generation behavior of LIFT-plus is relatively mild, with response lengths generally controlled within $200 \sim 400$. Its {\em Length Win Ratio} is only lower than \ourdpo{} by $33.62\%$ and $26.18\%$. However, for the {\em Base} models, LIFT-plus's response length skyrockets above $600$, causing severe semantic redundancy, and its {\em Length Win Ratio} significantly drops, being lower than \ourdpo{} by $39.53\%$ and $35.91\%$. This result, when analyzed alongside the findings in Section \ref{subsec:dpo-results} and Appendix~\ref{apx:short-bias-lift}, fully illustrates the drawbacks of LIFT-plus's over-reliance on length instructions and the complementary nature of \ourbtmodel{} in terms of both length instruction and semantic quality.

\begin{table*}[!ht]
    \centering
    \small
    \caption{Evaluation results of different DPO models on AlpacaEval-LI-plus-more.}
    \begin{tabular}{ccccccccc}
        \toprule
        \multirow{2}{*}{Metrics} & \multicolumn{4}{c}{Qwen2.5-7B-Base} & \multicolumn{4}{c}{Qwen2.5-7B-Instruct} \\
        \cmidrule(lr){2-5} \cmidrule(lr){6-9} & Baseline & LIFT-plus & R-DPO & Rc-DPO & Baseline & LIFT-plus & R-DPO & Rc-DPO \\
        \cmidrule(lr){1-1} Length Acc (\%) & 99.75 & \textbf{99.88} & 99.38 & \textbf{99.88} & 84.29 & 99.88 & 78.43 & \textbf{100} \\
        Response Length & 680.84 & 550.27 & 682.66 & 319.05 & 249.99 & 303.16 & 238.03 & 263.30 \\ 
        Length Win Ratio (\%) & 27.93 & 5.86 & 35.16 & \textbf{45.39} & 35.79 & 26.48 & 26.93 & \textbf{60.10} \\ 
        \midrule
        \multirow{2}{*}{Metrics} & \multicolumn{4}{c}{Llama-3.1-8B-Base} & \multicolumn{4}{c}{Llama-3.1-8B-Instruct} \\
        \cmidrule(lr){2-5} \cmidrule(lr){6-9} & Baseline & LIFT-plus & R-DPO & Rc-DPO & Baseline & LIFT-plus & R-DPO & Rc-DPO \\
        \cmidrule(lr){1-1} Length Acc (\%) & 89.03 & \textbf{99.75} & 93.39 & 97.51 & 98.25 & \textbf{99.88} & 97.01 & \textbf{99.88} \\
        Response Length & 448.77 & 664.20 & 540.34 & 429.69 & 279.04 & 413.42 & 246.73 & 279.56 \\ 
        Length Win Ratio (\%) & 36.03 & 9.73 & \textbf{47.76} & 45.64 & 45.14 & 33.67 & 47.01 & \textbf{59.85} \\
        \bottomrule
    \end{tabular}
    \label{tab:dpo-length-result-or-more}
\end{table*}

\subsection{Ablation Studies And Extra Results}
\label{appendix-ablation}
Due to the high cost of evaluating the DPO model, we primarily use the reward model for ablation studies. First, we conduct ablation studies to validate the key design choices of our approach. For convenience, we denote $\{(x, x_l^1, y_w)\}$ as $\mathcal{D}_{Rc}^c$ and $\{(x_l^2, x, y_l)\}$ as $\mathcal{D}_{Rc}^r$. 

\begin{figure}[!ht]
    \centering
    \vspace{-0.2cm}
    \subfigure[{\em w/o} $\mathcal{D}_{Rc}^r$]{\includegraphics[width=0.237\textwidth]{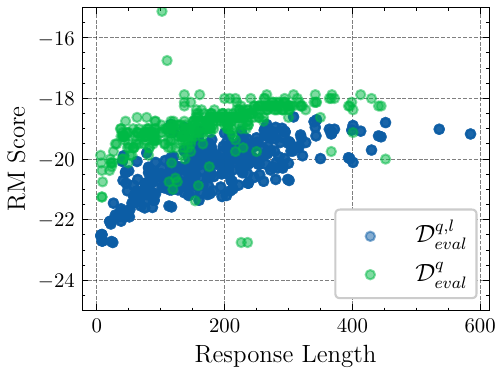}}
    \subfigure[{\em w/o} $\mathcal{D}_{Rc}^c$]{\includegraphics[width=0.237\textwidth]{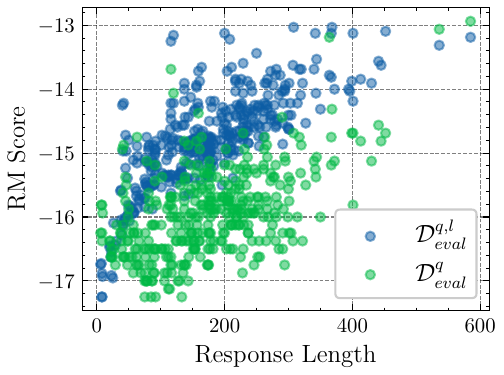}}
    \caption{The relationship between response length and RM score evaluated on $\mathcal{D}_{eval}^q$ and $\mathcal{D}_{eval}^{q,l}$ for Rc-RM trained with Llama-3.1-8B-Instruct.}
    \vspace{-0.2cm}
    \label{fig:llama-3.1-8b-instruct-origin-length-score-ranges-ablation}
\end{figure}

{\bf $\mathcal{D}_{Rc}^c$ and $\mathcal{D}_{Rc}^r$ are complementary.} We begin by proving the necessity of $\mathcal{D}_{Rc}^c$ and $\mathcal{D}_{Rc}^r$. To this end, we conduct ablation experiments using $\mathcal{D}_{rm} \cup \mathcal{D}_{Rc}^c$ ({\em w/o} $\mathcal{D}_{Rc}^r$) and $\mathcal{D}_{rm} \cup \mathcal{D}_{Rc}^r$ ({\em w/o} $\mathcal{D}_{Rc}^c$), with the results shown in Table \ref{tab:rm-result-ablation-Rc-BT}. As observed, when only one type of augmented dataset is used, the RM's {\em Quality Eval Acc} drops significantly, approaching the Baseline results in Table~\ref{tab:rm-result}. Simultaneously, its {\em Length Eval Acc} hovers around $50\%$, indicating that it fails to learn the length instruction. To further investigate, we prepend length instructions to each $x$ in $\mathcal{D}_{eval}^q$, forming $x_l$, where $y_w$ satisfies the length constraint in the instruction, , resulting in $\mathcal{D}_{eval}^{q,l}$. We then use the DPO models in Table \ref{tab:rm-result-ablation-Rc-BT} to score the pairs $(x_l, y_w)$ in $\mathcal{D}_{eval}^{q,l}$ and $(x, y_w)$ in $\mathcal{D}_{eval}^q$ to observe changes in the RM's predicted scores. In theory, for the pair $(x_l, y_w)$, since $y_w$ satisfies both the length constraint and original prompt of $x_l$, it should outperform $(x, y_w)$. That is, the score of reward model for $(x_l, y_w)$ should be higher than that for $(x, y_w)$. However, the results, illustrated in Figure \ref{fig:llama-3.1-8b-instruct-origin-length-score-ranges-ablation}(a), reveal that for the RM trained on $\mathcal{D} \cup \mathcal{D}_c$, the scores of $(x_l, y_w)$ with added length instructions are consistently lower than those of the original $(x, y_w)$, regardless of whether the added length constraint matches $y_w$'s length. This indicates that after training on $\mathcal{D}_{rm} \cup \mathcal{D}_{Rc}^c$, the RM learns a {\em length instruction bias}. Specifically, in $\mathcal{D}_{rm} \cup \mathcal{D}_{Rc}^c$, for any $(x_l, y_w)$, with an added length instruction, the reward model merely reduces the score compared to the original $(x, y_w)$ pair, without paying attention to the length instruction itself. A similar phenomenon in Figture~\ref{fig:llama-3.1-8b-instruct-origin-length-score-ranges-ablation} is observed in the RM trained on $\mathcal{D}_{rm} \cup \mathcal{D}_{Rc}^r$. Therefore, the combination of $\mathcal{D}_{Rc}^c$ and $\mathcal{D}_{Rc}^r$ is necessary to prevent the reward model from being exploited by this bias hacking.
\begin{figure*}[!ht]
    \centering
    \vskip 0.1in
    \subfigure[Qwen2-1.5B-Instruct]{\includegraphics[width=0.4\textwidth]{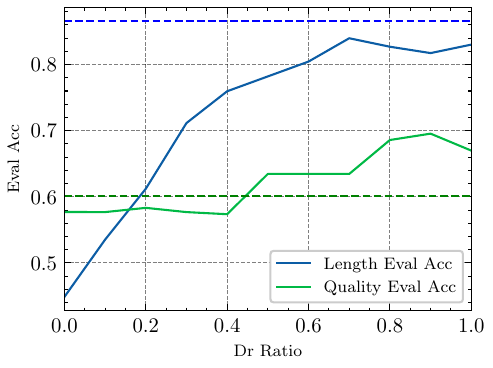}}
    \subfigure[Llama-3.1-8B-Instruct]{\includegraphics[width=0.4\textwidth]{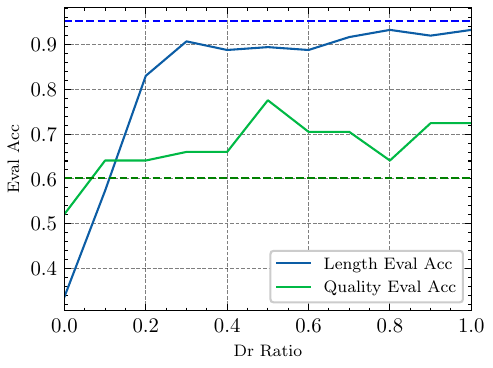}}
    \caption{Variations in Rc-RM's {\em Length Eval Acc} and {\em Quality Eval Acc} as $\mathcal{D}_{Rc}^r$ increases. The green dashed line denotes the Baseline {\em Quality Eval Acc}, while the blue dashed line represents the {\em Length Eval Acc} of LIFT-plus.}
    \label{fig:llama-3.1-8b-instruct-ablation-Dr-range-rm-score}
    \vskip -0.1in
\end{figure*}

\begin{table*}[!ht]
    \centering
    \caption{Evaluation results of different reward models on quality ($\mathcal{D}_{eval}^q$) and length ($\mathcal{D}_{eval}^l$) evaluation datasets (HH-RLHF).}
    \vskip 0.1in
    \begin{tabular}{llcc}
    \toprule
        Model & Variant & Quality Eval Acc ($\%$) & Length Eval Acc ($\%$) \\ 
        \midrule
        \multirow{4}{*}{Qwen2-1.5B-Instruct} & Baseline & 58.65 & 59.36 \\ 
        & LIFT-plus & 58.01 & \textbf{85.90} \\ 
        & ODIN & 59.62 & 48.72 \\
        & Rc-RM & \textbf{61.22} & 85.26 \\ 
        \midrule
        \multirow{4}{*}{Qwen2.5-7B-Instruct} & Baseline & 62.50 & 52.24 \\ 
        & LIFT-plus & 57.37 & \textbf{92.63} \\ 
        & ODIN & 65.06 & 57.69 \\
        & Rc-RM & \textbf{65.38} & 91.03 \\ 
        \midrule
        \multirow{4}{*}{Llama-3.1-8B-Instruct} & Baseline & 63.14 & 51.60 \\ 
        & LIFT-plus & 58.33 & \textbf{93.27} \\ 
        & ODIN & 64.10 & 50.00 \\
        & Rc-RM & \textbf{69.23} & 91.03 \\ 
        \bottomrule
    \end{tabular}
    \vskip -0.1in
    \label{tab:rm-result-hh}
\end{table*}

{\bf Effectiveness of different ratios of $\mathcal{D}_{Rc}$.} We conducted further experiments to demonstrate the efficiency of the augmented dataset $\mathcal{D}_{Rc}$ we constructed. Specifically, we gradually increased $\mathcal{D}_{Rc}$ from $0\%$ to $100\%$ on the original training dataset $\mathcal{D}_{rm}$, with a $10\%$ increment, to train the reward models for {\em Qwen2-1.5B-Instruct} and {\em Qwen2.5-7B-Instruct}. The results of {\em Quality Eval Acc} (\%) are shown in Table \ref{tab:rm-result-D_Rc_ratio}. It can be observed that our method demonstrates superlinear performance improvements with a smaller proportion of $\mathcal{D}_{Rc}$. In other words, using only $40\%$ of the augmented data yields approximately a $70\%$ performance improvement. This indicates that, even under tight computational constraints, our method can achieve significant performance gains with the addition of a small amount of data, which fully demonstrates the high efficiency of our augmented dataset $\mathcal{D}_{Rc}$ and our method.

\begin{table*}[!ht]
    \centering
    \small
    \caption{\QEA{} of different reward models on quality ($\mathcal{D}_{eval}^q$) evaluation datasets under different proportions of $\mathcal{D}_{Rc}$.}
    \begin{tabular}{lccccccccccc}
    \toprule
        Model & 0\% & 10\% & 20\% & 30\% & 40\% & 50\% & 60\% & 70\% & 80\% & 90\% & 100\% \\ 
        \midrule
        Qwen2-1.5B-Instruct & 60.75 & 64.43 & 65.69 & 66.68 & 67.31 & 68.11 & 68.77 & 69.95 & 70.40 & 71.05 & 71.47 \\ 
        \midrule
        Qwen2.5-7B-Instruct & 59.31 & 61.92 & 64.26 & 66.33 & 68.12 & 69.63 & 70.87 & 71.83 & 72.52 & 72.93 & 73.07 \\  
        \bottomrule
    \end{tabular}
    \label{tab:rm-result-D_Rc_ratio}
\end{table*}

{\bf Effectiveness of different ratios of $\mathcal{D}_{Rc}^c$ and $\mathcal{D}_{Rc}^r$ mixtures.} We also briefly explore the impact of different ratios of $\mathcal{D}_{Rc}^c$ and $\mathcal{D}_{Rc}^r$ on the final RM results. Specifically, we fix $\mathcal{D}_{rm}$ and $\mathcal{D}_{Rc}^c$ constant, and increase $\mathcal{D}_{Rc}^r$ from $0\%$ to $100\%$ with a $10\%$ increment. The results are shown in Figure \ref{fig:llama-3.1-8b-instruct-ablation-Dr-range-rm-score} (a) and (b). As observed, initially, the model is biased toward the one-sided data of $\mathcal{D}_{Rc}^c$, leading to low {\em Quality Eval Acc} and {\em Length Eval Acc}, even falling below the Baseline models (denoted as green dashed line). As $\mathcal{D}_{Rc}^r$ increased, {\em Length Eval Acc} rises rapidly, with the growth rate slowing down around $30\%$, while {\em Quality Eval Acc} gradually increases.
It effectively demonstrates the complementary effect of $\mathcal{D}_{Rc}^c$ and $\mathcal{D}_{Rc}^r$ in mitigating length bias and enhancing adherence to length instructions.

{\bf Effectiveness across different datasets.} To further validate the generalizability of \ourbtmodel{}, we conduct RM training on the HH-RLHF \citep{bai2022training} dataset\footnote{We use {\em Instruct} models for direct RM training on the HH-RLHF dataset.}, while still using $\mathcal{D}_{eval}^q$ and $\mathcal{D}_{eval}^l$ for evaluation. The results are shown in Table \ref{tab:rm-result-hh}. As seen, despite the HH-RLHF dataset being from a different domain compared to the OpenAssistant dataset \citep{kopf2024openassistant}, which can be considered an out-of-distribution (OOD) scenario, \ourrm{} still significantly outperforms other methods on $\mathcal{D}_{eval}^q$, while achieving results similar to LIFT-plus on $\mathcal{D}_{eval}^l$. Specifically, for Qwen2-1.5B-Instruct, \ourrm{} outperforms Baseline by $2.57\%$ and ODIN by $1.60\%$. For Qwen2.5-7B-Instruct, \ourrm{} surpasses Baseline by $2.88\%$ and ODIN by $0.32\%$. Notably, for Llama-3.1-8B-Instruct, \ourrm{} significantly outperforms Baseline by $6.09\%$ and also surpasses ODIN by $5.13\%$, further demonstrating the effectiveness of \ourrm{} in mitigating length bias. Moreover. on $\mathcal{D}_{eval}^l$, \ourrm{} achieves results comparable to LIFT, reaffirming its effectiveness in following length instructions. Finally, the consistent performance across different datasets provides strong evidence of the robustness and effectiveness of \ourbtmodel{} in diverse scenarios.

{\bf Effectiveness across different bias.} We used format bias as another test bed, as it is the another common type of bias observed in LLMs. We used the FormatBiasEval dataset \citep{long2024llms}, which consists of multiple-choice questions with specific formatting requirements, to train and evaluate reward models. As the dataset comes with a predefined split into training and evaluation subsets, we adhere to this split in our experiments. To adapt the training subset into a preference dataset, we designated the correct option as the chosen response and randomly selected incorrect option as the rejected response. Then we constructed additional "response-conditioned" triples of the form $(x, x_f^1, y_w)$ and $(x, x_f^2, y_w)$, where $y_w$ violates the format constraints in $x_f^1$ (or $x_f^2$) while $y_l$ satisfies it. The final augmented dataset is defined as $\mathcal{D}_{\text{rm}}^f = \{(x, x_f^1, y_w)\} \cup \{(x, x_f^2, y_w)\}$.

Due to the programmatically constructible nature of the format, we created a semantic evaluation dataset based on the evaluation subset, referred to as $\mathcal{D}_{eval}^{f}$, using a methodology similar to that used for constructing $\mathcal{D}_{eval}^q$, but without relying on GPT-4o. The process is as follows: For each triple $(x, y_w, y_l)$ in the original evaluation dataset, we generated two complementary triples: $(x, f(y_w), y_l)$ and $(x, y_w, f(y_l))$, where $f(y)$ denotes a program that applies a randomly selected format pattern to the response $y$. Therefore, the $D_{eval}^{q,f}$ can effectively mitigate the impact of format bias and facilitate more fair evaluation of the reward model's semantic capabilities.

\begin{table}[!ht]
    \centering
    \small
    \caption{The accuracy of different reward models on quality ($\mathcal{D}_{eval}^{f}$) evaluation datasets.}
    \label{tab:rm-result-format}
    \setlength{\tabcolsep}{5pt}
    \begin{tabular}{lccc}
        \toprule
        Model & Baseline & Pc-Format & Rc-RM \\
        \midrule
        Qwen2-1.5B-Instruct & 76.25 & 75.41 & \textbf{79.17} \\
        Qwen2.5-7B-Instruct & 82.42 & 78.33 & \textbf{89.27} \\
        Llama-3.1-8B-Instruct & 79.39 & 79.01 & \textbf{84.24} \\
        \bottomrule
    \end{tabular}
    \vspace{-0.2cm}
\end{table}

We conduct experiments on {\em Instruct} models. And the final results are shown in Table \ref{tab:rm-result-format}, where we refer to the prompt-conditioned method, which is similar to LIFT, as {\bf Pc-Format}. As shown, \ourrm{} remains more effective in mitigating format bias, achieving superior accuracy across all evaluated models. These results clearly highlight the strong generalization ability of our method.

\begin{figure*}[!ht]
    \centering
    \includegraphics[width=0.8\textwidth]{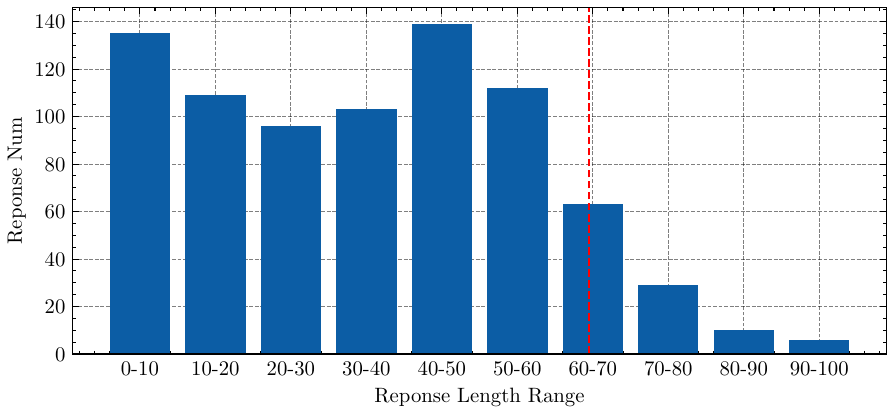}
    \caption{Response length range generated by LIFT-plus trained on Llama-3.1-8B-Instruct using length instruction $x_l$ in AlpacaEval-LI-plus-less benchmark. The red dashed line $x = 64$ represents the $90\%$ threshold, indicating that $90\%$ of the responses have a length less than or equal to $64$.}
    \label{fig:response_length-range-lift-length-instruct}
\end{figure*}

\subsection{The "Short Bias" in LIFT-plus}
\label{apx:short-bias-lift}
From the perspective of {\em Response Length} in Table~\ref{tab:dpo-origin-quality-result}, it might appear that LIFT-plus exhibits smaller length bias. This is not the case. Our results demonstrate that while LIFT-plus reduces response length, it does so at the expense of disregarding the balance between response length and semantic quality, resulting in suboptimal performance in both semantic quality and adherence to length instructions. Specifically, our results indicate that DPO models trained with the LIFT-plus method tend to exhibit a "short bias" phenomenon, where the model prioritizes generating shorter responses without considering semantic quality, given the "or less" instruction $x_l$ or the original prompt $x$. As shown in Table \ref{tab:dpo-origin-quality-result} and \ref{tab:dpo-length-result-or-less}, both for {\em Quality Win Ratio} and {\em Length Win Ratio}, LIFT-plus significantly underperforms compared to R-DPO and \ourbtmodel{}. Notably, for Qwen2.5-7B-Base and Qwen2.5-7B-Instruct, LIFT-plus's {\em Quality Win Ratio} is lower than R-DPO by $8.73\%$ and $8.47\%$, and lower than \ourbtmodel{} by $13.72\%$ and $18.94\%$, respectively. A similar trend is observed with Llama-3.1-8B-Base and Llama-3.1-8B-Instruct. Furthermore, we plot the distribution of response lengths generated by LIFT-plus trained on Llama-3.1-8B-Instruct under the length instruction $x_l$ in AlpacaEval-LI-plus-less benchmark, as shown in Figure \ref{fig:response_length-range-lift-length-instruct}. As observed, the mean response length specified by the length instruction $x_l$ in AlpacaEval-LI-plus-less is $180.23$, while the response lengths generated by LIFT-plus are all below $100$. Furthermore, more than $90\%$ of the responses have length under $64$. This clearly demonstrates that the DPO models trained with LIFT-plus focus solely on the "or less" length instruction $x_l$, disregarding the semantic requirements of the original prompt $x$ and even overlooking the specific length constraint ({\em word\_num}) of the $x_l$ itself, resulting in the "short bias".

\end{document}